\documentclass{article}
\usepackage[preprint, nonatbib]{arxiv}

\usepackage[utf8]{inputenc}
\usepackage[T1]{fontenc}
\usepackage{hyperref}
\usepackage{url}
\usepackage{booktabs}
\usepackage{amsmath,amssymb,amsfonts,amsthm,mathrsfs}
\usepackage{graphicx}
\usepackage{multirow}
\usepackage[table]{xcolor}
\usepackage{textcomp}
\usepackage{array,longtable,tabularx,ltablex,ragged2e}
\usepackage{pdflscape}
\usepackage[title]{appendix}
\usepackage{adjustbox}
\usepackage{caption}
\usepackage{subcaption}
\usepackage{threeparttable}
\usepackage{float}
\usepackage{placeins}
\usepackage[most]{tcolorbox}
\usepackage{listings}
\usepackage{upquote}
\usepackage{makecell}
\usepackage{microtype}
\usepackage{newtxtext,newtxmath}
\usepackage{enumitem}
\hypersetup{colorlinks=true,linkcolor=black,citecolor=blue,urlcolor=blue}
\graphicspath{{./}{figs/}}
\setlist{nosep,leftmargin=*}

\AtBeginDocument{
\newgeometry{
  textheight=9.50in,
  textwidth=6.50in,
  top=0.5in,
  headheight=12pt,
  headsep=16pt,
  footskip=32pt
}}

\lstdefinestyle{promptmono}{
  float,
  floatplacement=t,
  upquote=true,
  literate=
    {“}{"}1 {”}{"}1 {„}{"}1 {‟}{"}1%
    {‘}{'}1 {’}{'}1 {‛}{'}1%
    {−}{-}1 {–}{-}1 {—}{-}1 {‒}{-}1%
    {…}{...}3,
  basicstyle=\small,
  backgroundcolor=\color{gray!3},
  frame=single,
  rulecolor=\color{black!70},
  framesep=3pt,
  frameround=tttt,
  columns=fullflexible, keepspaces=false,xleftmargin=0pt,
  breaklines=true, showstringspaces=false, tabsize=2
}

\definecolor{CaseHeader}{HTML}{FFF2CC}
\definecolor{ClinInit}{HTML}{E2EFDA}
\definecolor{ModelCoT}{HTML}{F2F2F2}
\definecolor{Revised}{HTML}{DDEBF7}
\newtcolorbox{bandbox}[1]{enhanced,breakable,
  colback=#1, colframe=white, boxrule=0pt,
  left=2pt, right=2pt, top=2pt, bottom=2pt, arc=1.5mm,
  before skip=2pt, after skip=2pt
}

\title{Towards Precision Therapy in Hepatocellular Carcinoma: A Clinical-Reasoning LLM for Risk Stratification and Treatment Guidance}

\author{%
\textbf{Peng Cui}\textsuperscript{1,*}\quad \textbf{Jitao Wang}\textsuperscript{2,*}\quad \textbf{Siyan Xue}\textsuperscript{3,*}\quad \textbf{Yao Huang}\textsuperscript{4,*}\quad \textbf{Haoming Xia}\textsuperscript{2,*}\quad
\textbf{Dong Li}\textsuperscript{2}\quad \textbf{Dengxiang Liu}\textsuperscript{5}\\ \textbf{Weilin Wang}\textsuperscript{6}\quad \textbf{Liping Liu}\textsuperscript{7,8}\quad \textbf{Leida Zhang}\textsuperscript{9}\quad
\textbf{Yunfu Cui}\textsuperscript{10}\quad \textbf{Tao Peng}\textsuperscript{11}\quad \textbf{Daolin Ji}\textsuperscript{12}\quad \textbf{Haitao Zhao}\textsuperscript{13} \\ \textbf{Wei Zhang}\textsuperscript{14}\quad
\textbf{Xiaojuan Wang}\textsuperscript{2}\quad \textbf{Weijie Ma}\textsuperscript{15,16}\quad \textbf{Zongren Ding}\textsuperscript{17}\quad \textbf{Jinlong Li}\textsuperscript{5}\quad \textbf{Yuan Ding}\textsuperscript{6}\quad
\textbf{Jiajing Zhao}\textsuperscript{7,8}\quad
\textbf{Zhiyu Chen}\textsuperscript{9}\quad \textbf{Chengkun Yang}\textsuperscript{11}\quad \textbf{Ziyue Huang}\textsuperscript{13}\quad \textbf{Jiaqi Liu}\textsuperscript{2}\quad
\textbf{Fusheng Liu}\textsuperscript{15,16}\quad \textbf{Yang Zhou}\textsuperscript{18}\quad \textbf{Xiaojuan Wang}\textsuperscript{5}\quad \textbf{Zhongquan Sun}\textsuperscript{6}\quad \textbf{Shiyun Bao}\textsuperscript{7,8}\quad
\textbf{Xiaojun Wang}\textsuperscript{9}\quad \textbf{Ming Yang}\textsuperscript{2}\quad \textbf{Guangxin Li}\textsuperscript{19}\quad \textbf{Bin Shu}\textsuperscript{2}\quad \textbf{Yong Liao}\textsuperscript{2}\quad
\textbf{Hongxuan Li}\textsuperscript{2}\quad \textbf{Yao Tang}\textsuperscript{2}\quad \textbf{Shizhong Yang}\textsuperscript{2}\quad \textbf{Yongyi Zeng}\textsuperscript{17}\quad \textbf{Yufeng Yuan}\textsuperscript{15,16}\quad
\textbf{Yinpeng Dong}\textsuperscript{4,\textdagger}\quad \textbf{Jihui Hao}\textsuperscript{20,\textdagger}\quad \textbf{Jun Zhu}\textsuperscript{1,\textdagger}\quad \textbf{Jiahong Dong}\textsuperscript{2,\textdagger}\\[0.9em]
\small \textsuperscript{1}Department of Computer Science \& Technology, Tsinghua University, Beijing, China\\
\small \textsuperscript{2}Hepato-Pancreato-Biliary Center, Beijing Tsinghua Changgung Hospital, Key Laboratory of Digital Intelligence Hepatology (Ministry of Education), School of Clinical Medicine, Tsinghua Medicine, Tsinghua University, Beijing, China\\
\small \textsuperscript{3}School of Biomedical Engineering, Tsinghua University, Beijing, China\\
\small \textsuperscript{4}College of AI, Tsinghua University, Beijing, China\\
\small \textsuperscript{5}Hebei Provincial Key Laboratory of Portal Hypertension \& Cirrhosis, Xingtai People's Hospital, Xingtai, China\\
\small \textsuperscript{6}Department of Hepatobiliary and Pancreatic Surgery, The Second Affiliated Hospital, Zhejiang University School of Medicine, Hangzhou, China\\
\small \textsuperscript{7}Division of Hepatobiliary and Pancreas Surgery, Department of General Surgery, Shenzhen People's Hospital (The First Affiliated Hospital, Southern University of Science and Technology, The Second Clinical Medical College, Jinan University), Shenzhen, China\\
\small \textsuperscript{8}The Second Clinical Medical College, Jinan University, Shenzhen, China\\
\small \textsuperscript{9}Department of Hepatobiliary Surgery, Southwest Hospital, Third Military Medical University (Army Medical University), Chongqing, China\\
\small \textsuperscript{10}Department of Hepatopancreatobiliary Surgery, Second Affiliated Hospital of Harbin Medical University, Harbin, China\\
\small \textsuperscript{11}Department of Hepatobiliary Surgery, The First Affiliated Hospital of Guangxi Medical University, Nanning, China\\
\small \textsuperscript{12}Department of Hepatopancreatobiliary Surgery, The Fourth Affiliated Hospital, Harbin Medical University, Harbin, China\\
\small \textsuperscript{13}Department of Liver Surgery, Peking Union Medical College Hospital, Chinese Academy of Medical Sciences and Peking Union Medical College (CAMS \& PUMC), Beijing, China\\
\small \textsuperscript{14}Department of Hepatobiliary Surgery, Tianjin Medical University Cancer Institute and Hospital, National Clinical Research Center for Cancer, Tianjin Key Laboratory of Digestive Cancer, Tianjin's Clinical Research Center for Cancer, Tianjin, China\\
\small \textsuperscript{15}Department of Hepatobiliary \& Pancreatic Surgery, Zhongnan Hospital of Wuhan University, Wuhan, China\\
\small \textsuperscript{16}Clinical Medicine Research Center for Minimally Invasive Procedure, Wuhan University, Wuhan, China\\
\small \textsuperscript{17}Department of Hepatopancreatobiliary Surgery, Mengchao Hepatobiliary Hospital of Fujian Medical University, Fuzhou, China\\
\small \textsuperscript{18}Biological Information Biobank, Mengchao Hepatobiliary Hospital of Fujian Medical University, Fuzhou, China\\
\small \textsuperscript{19}Department of Radiotherapy, Beijing Tsinghua Changgung Hospital, School of Clinical Medicine, Tsinghua Medicine, Tsinghua University, Beijing, China\\
\small \textsuperscript{20}Pancreas Center, Tianjin Medical University Cancer Institute and Hospital, National Clinical Research Center for Cancer, State Key Laboratory of Druggability Evaluation and Systematic Translational Medicine, Tianjin Key Laboratory of Digestive Cancer, Tianjin’s Clinical Research Center for Cancer, Tianjin, China\\[0.7em]
\small \textsuperscript{*}These authors contributed equally to this work.\quad \textsuperscript{\textdagger}Corresponding authors: dongyinpeng@tsinghua.edu.cn, haojihui@tjmuch.com, dcszj@tsinghua.edu.cn, dongjiahong@mail.tsinghua.edu.cn
}

\begin{document}

\maketitle
\newpage
\begin{abstract}
Hepatocellular carcinoma (HCC) ranks among the most prevalent malignancies worldwide and remains a leading cause of cancer-related mortality. However, prevailing guidelines and staging systems delineate only coarse categories, failing to capture within-stage heterogeneity and overlooking the clinical context embedded in electronic medical records (EMRs). To bridge this gap, we present \textit{HCC-STAR} (Hepatocellular Carcinoma Staging, Treatment And pRognosis), a clinically aligned large language model that reads routine EMR narratives and jointly outputs a risk score-based staging, a ranked list of guideline-consistent treatments with evidence-based rationales, and individualized survival estimates. We curate approximately 30{,}000 HCC cases from SEER and expand them into EMR-style narrative training data via a clinician-validated, prompt-based augmentation workflow. Building on this corpus, we propose a knowledge-aligned, reasoning-centric training framework, optimized under a step-verifiable composite reward, that departs from text-level memorization of clinical guidelines.
Evaluated on the multi-center cohort of 6{,}668 patients from 12 hospitals across China, HCC-STAR achieves state-of-the-art performance in treatment recommendation and risk stratification compared with clinical guidelines and competitive models, particularly GPT-5 and Gemini-2.5 Pro. Hypothetical overall-survival analyses indicate that the median survival under adherence to the model's recommendations was 51 months, compared with 29 and 32 months under BCLC and CNLC, respectively. In a clinician-centric evaluation, blinded hepatobiliary specialists rate HCC-STAR’s chain of thought and evidence-based justifications as trustworthy. The model surpasses resident and attending physicians in treatment accuracy and, when used as an assistant, helps them make more accurate decisions faster.
These findings support HCC-STAR as a reliable and verifiable decision-support system for risk stratification and precision therapy in HCC.
\end{abstract}

\textbf{Keywords:} Hepatocellular carcinoma; large language model; reinforcement learning; precision therapy; survival prognosis

\section{Main}\label{sec1}
Hepatocellular carcinoma (HCC) is the sixth most common malignancy worldwide and the third leading cause of cancer-related mortality, with particularly high incidence in East Asia and sub-Saharan Africa~\cite{chan2025lancet,bray2024global}. Despite advances in screening, surgical technique, and systemic therapy, overall outcomes remain poor owing to delayed or suboptimal treatment and the inherent heterogeneity of tumor biology and patient conditions~\cite{safri2024heterogeneity,rich2020hepatocellular}. Analyses of large real-world cohorts indicate that a substantial proportion of patients experience treatment delays exceeding 90 days from diagnosis, which are strongly associated with reduced survival, particularly in early-stage disease~\cite{cheo2024impact,govalan2022therapeutic}. Clinical decision-making in HCC is inherently complex: clinicians must synthesize information on tumor burden, liver function, performance status, vascular invasion, and extrahepatic spread across multiple modalities. In practice, such decisions frequently depend on a small number of senior hepatobiliary specialists and iterative multidisciplinary discussions within constrained time frames. Limited access to expert consultation and uneven healthcare resources contribute to missed opportunities for timely, individualized therapy, resulting in suboptimal real-world outcomes.

Over the course of decades of progress in hepatology, consensus guidelines and staging systems, including AJCC (TNM)~\cite{ajcc8_manual}, BCLC~\cite{Reig2022BCLC}, and CNLC~\cite{Zhou2023CNLC}, have established structured principles for treatment selection by linking staging systems to preferred treatment pathways. Nevertheless, these frameworks were primarily designed for population-level management and insufficiently accommodate patient-level heterogeneity, such as marginal hepatic reserve, multifocal disease, comorbidities, or nuanced imaging phenotypes~\cite{llovet_hepatocellular_2021}. As a result, they lack the flexibility to combine individualized treatment recommendations with prognostic estimation within a single workflow. Moreover, mastering guideline intricacies and accumulating case-based expertise require extensive training and experience, hindering consistent, high-quality decision-making across diverse healthcare settings. In HBV-predominant cohorts, refined risk stratification further delineates subgroups that derive greater benefit from curative or intensified interventions, underscoring the limitations of guideline-based staging alone in achieving personalized care~\cite{yau2014development}. Collectively, these challenges highlight the need for clinically grounded artificial intelligence that integrates fine-grained clinical variables with risk stratification and treatment decision-making at the level of individual patients.

Recent advances in large language models (LLMs)~\cite{Brown2020GPT3,OpenAI2023GPT4} and reasoning-enhanced training paradigms~\cite{ouyang2022training,wei2022chain,shao_deepseekmath_2024} offer promising avenues for clinical decision support. However, most existing works and remarkable performance remain diagnosis-centric~\cite{BioMedLM,BioGPT,MedPaLM,MedPaLM2,liu2025generalist}, facilitated by well-defined criteria and curated datasets, whereas comparatively few address \emph{treatment recommendation} or \emph{longitudinal decision-making} in oncology. High scores on diagnostic benchmarks can not guarantee robust generalization to heterogeneous real-world cohorts for treatment recommendations or risk stratifications because treatment planning is a multi-step process that must integrate diverse clinical data and be continually updated. Limited cross-institutional validation and the absence of oncology-specific adaptation further constrain clinical reliability~\cite{zhu2025large}. Recent studies~\cite{qiu2025quantifying,hager2024evaluation} also show that, even when state-of-the-art LLMs achieve high accuracy on diagnostic tasks and medical licensing exams (e.g., PubMedQA~\cite{jin2019pubmedqa}, USMLE~\cite{kung2023performance}), their performance on treatment planning remains substantially lower, and guideline-concordant recommendations are often unreliable in realistic clinical workflows. 
Beyond diagnosis, a few studies have begun to explore disease management and longitudinal care by leveraging LLMs and multimodal vision–language systems for primary diabetes management and retinal disease screening~\cite{li2024integrated,wu2025eyecare}. For HCC, while prior work~\cite{HEROVision} attempts to optimize treatment (ablation and resection) with a Vision Transformer~\cite{DosovitskiyB0WZ21} for recurrent HCC meeting Milan criteria, its imaging-only design and limited decision space fall short of the broader aims of this paper, i.e., guideline-aligned treatment recommendation, fine-grained staging guidance, and survival prediction.

\begin{figure}[t]
    \centering
    \includegraphics[width=0.95\linewidth,height=0.68\textheight,keepaspectratio]{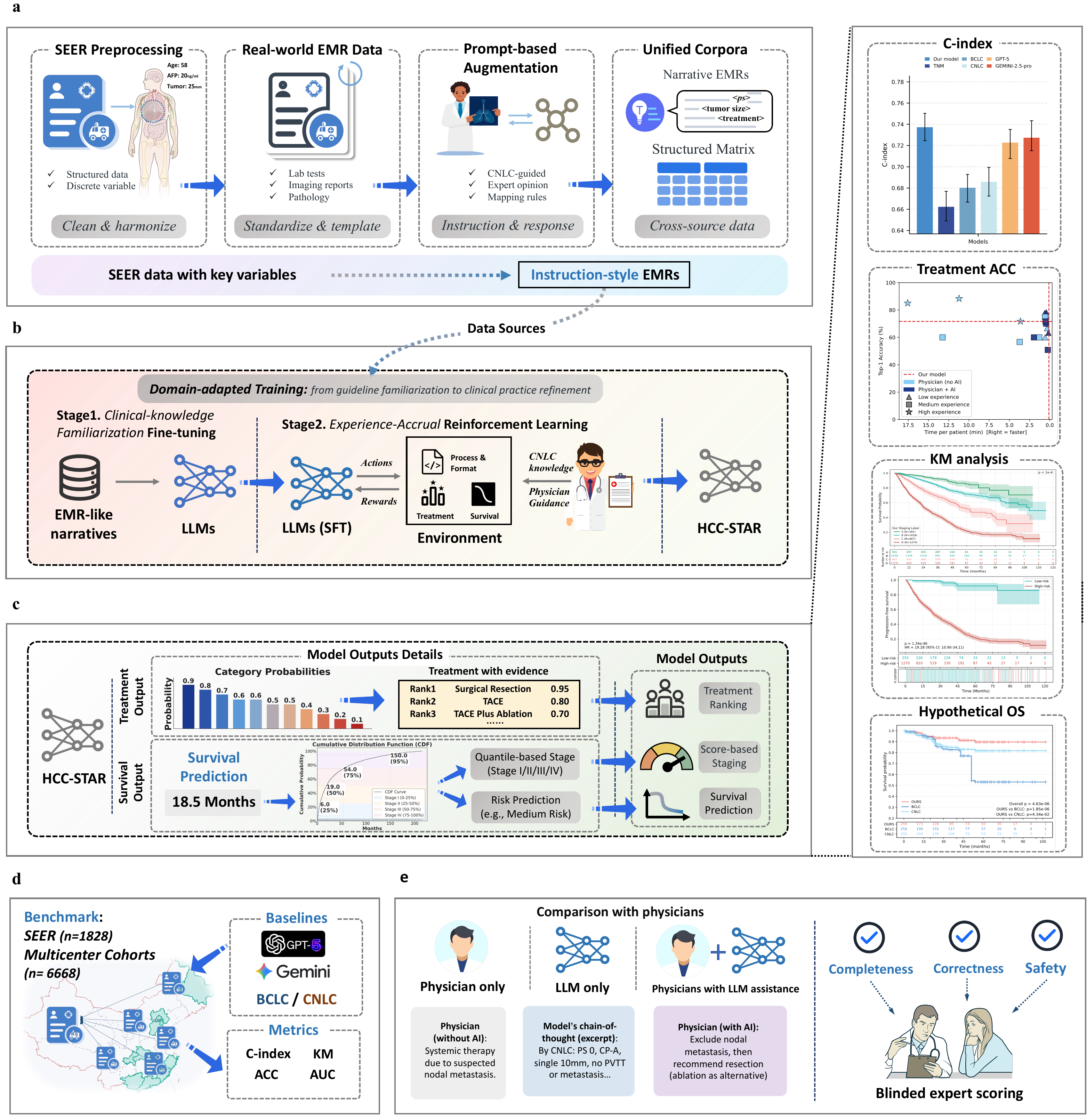}
    \caption{Overview of HCC-STAR: data curation, model development, and evaluation.
    \textbf{a}, {Data acquisition, augmentation, and harmonization.} HCC records from SEER are cleaned and normalized, and real-world EMRs are standardized into templates. A CNLC-guided, clinical-validated prompt pipeline expands key variables into instruction-style EMR narratives. Structured matrices and narratives are then unified into a cross-source corpus.
    \textbf{b}, {Model training.} Stage 1: Clinical-Knowledge Familiarization Fine-Tuning on EMR-like narratives. Stage 2: Experience-Accrual Reinforcement Learning with a clinically tailored composite reward (process/format correctness, guideline-consistent treatment ranking, survival estimation, brevity, and chain of thought (CoT) quality).
    \textbf{c}, {Model outputs.} Given an EMR-style input, HCC-STAR returns (i) category probabilities and a ranked treatment list with guideline-linked rationales, (ii) a risk score-based stage via quantile calibration, and (iii) a patient-specific survival estimate (months) together with a risk tier.
    \textbf{d}, {Model validation.} Internal (SEER) and external multi-center cohorts benchmark our model against staging systems (AJCC/TNM, BCLC, CNLC), representative LLMs, and classical ML using Top-$k$ accuracy, Harrell’s C-index, AUROC, and Kaplan–Meier analyses.
    \textbf{e}, {Clinical comparison and evaluation of LLM-generated content.} We conduct evaluations with physicians and compare three settings: physician only, LLM only, and physicians with LLM assistance. Blinded hepatobiliary specialists score the CoT for completeness, correctness, safety and the reliability of evidence-based justifications.}
    \label{fig:overview}
\end{figure}

To address these limitations, we introduce \textit{HCC-STAR} (Hepatocellular Carcinoma Staging, Treatment And pRognosis), a domain-adapted model that integrates fine-grained HCC staging, individualized treatment recommendation, and survival prediction within a unified framework (Fig.~\ref{fig:overview}c). For model development, we first curate approximately 30{,}000 HCC cases with prognostic annotations from Surveillance, Epidemiology,
and End Results (SEER) program~\cite{seer_overview} as the primary training corpus. However, SEER contains only a limited set of structured variables, such as age, tumor size, alpha-fetoprotein (AFP), TNM stage, and survival months, but lacks the rich textual detail required for supervised fine-tuning and reasoning-oriented learning in LLMs. To bridge this gap, we design a prompt-based data augmentation pipeline informed by distilled clinical knowledge, including treatment guidelines and evidence-based justifications, derived from both the Chinese Expert Consensus on Conversion and Perioperative Therapy of Primary Liver Cancer (CNLC 2024)~\cite{jia2024chinese} and de-identified real-world electronic medical records (EMRs) (Fig.~\ref{fig:overview}a). Through this process, LLMs expand structured entries from SEER into coherent EMR-style narratives, yielding a large-scale and clinically realistic corpus. This strategy compensates for missing clinical context and improves the reliability of synthetic data, establishing a robust foundation for model training.

We adopt a two-stage training paradigm that implements a \emph{knowledge-aligned reasoning paradigm} rather than text-level guideline memorization (Fig.~\ref{fig:overview}b). In Stage~1, \emph{Clinical-Knowledge Familiarization Fine-Tuning (CKF-FT)} guides the model to behave as a hepatobiliary clinician by working through \emph{worked clinical examples} using EMR-style narratives augmented with guideline logic and evidence citations. In this way, guideline-consistent reasoning and clinically grounded outputs are learned through practice rather than verbatim copying. In Stage 2, \emph{Experience-Accrual Reinforcement Learning (EARL)} refines clinical reasoning under a clinically tailored composite reward with two central design choices. First, the reward is step-verifiable: it inspects intermediate variables in the chain of thought (CoT), including staging cues, performance status, Child–Pugh grade, vascular invasion and evidence citations, rather than scoring only the final answer, so that the model is rewarded for arriving at the right answer for the right reason. Second, treatment-ranking and survival-estimation rewards are decoupled during Group Relative Policy Optimization (GRPO)~\cite{shao_deepseekmath_2024}, preventing gradient interference between ranking supervision and censored-time supervision—two objectives that otherwise pull the policy in different directions. Together, these choices convert guideline logic into an auditable, multi-objective and stable learning signal. Given that guideline logic is exercised during reasoning rather than memorized, incremental guideline updates can be incorporated via system-prompt edits without model retraining. Across both stages, the framework ingests heterogeneous clinical narratives end-to-end (diagnostic notes, imaging reports, operative summaries, discharge documentation), encouraging the model to surface salient factors and link reasoning steps to guideline-aligned actions. The multi-objective reasoning framework yields the CoT~\cite{cot} that is accurate, interpretable, and clinically faithful, enabling HCC-STAR to adhere to established standards while iteratively improving treatment precision and individualized risk stratification.

Across retrospective cohorts and blinded clinical evaluations, HCC-STAR consistently outperforms traditional machine-learning baselines, representative large language models (GPT-5~\cite{openai_gpt5_2025}, Gemini-2.5-Pro~\cite{google_gemini25_2025}, GPT-4o~\cite{openai_gpt4o_syscard}, Claude~\cite{anthropic_claude35_2024}, DeepSeek-R1~\cite{deepseek_r1_2025}), and major guideline-based staging systems (AJCC/TNM, BCLC, CNLC) for both treatment accuracy and survival prediction. In hypothetical overall-survival analyses, adherence to the model’s recommendations is associated with substantially longer survival than following BCLC/CNLC (median OS 51 vs 29–32 months), highlighting potential clinical utility. Unlike most ML models trained on fixed and structured schemas that often fail to generalize across heterogeneous data formats, our approach interprets EMR-style narratives, captures nonlinear clinical interactions, and generates explicit CoT rationales, enabling cross-institutional generalization without retraining. In external multi-center validation, the model achieves the highest C-index and AUROC, demonstrating the benefit of continuous risk estimation that captures within-stage heterogeneity. In physician-level comparisons, HCC-STAR surpasses junior and intermediate clinicians in treatment accuracy and approaches the performance of senior specialists. When used as an assistive tool, it significantly improves the accuracy and efficiency of less experienced physicians, narrowing disparities in decision quality. In a structured evaluation led by hepatobiliary experts, our model's CoT reasoning is rated more complete, accurate, and safer than that of GPT-4o and DeepSeek-R1, and with more precise and reliable evidence-based justification. Taken together, these results position HCC-STAR as, to our knowledge, the first domain-adapted reasoning model that unifies risk stratification and treatment decision-making in HCC, underscoring its potential to advance LLM-assisted clinical decision-making in hepatology and catalyze broader exploration of reasoning-centric LLMs in disease therapy and prognostication.

\section{Results}\label{sec2}
\subsection{Data characteristics and model development}
We first characterized the internal SEER cohort and the external multicenter HCC cohort used for model development and evaluation. The SEER Program contributed \mbox{$\sim$30{,}000} structured HCC cases (diagnosed 2004--2020)  for internal training (\mbox{$\sim$28{,}000}) and testing (\mbox{$\sim$2{,}000}). In parallel, we assembled a real-world external cohort of 6,668 patients with hepatocellular carcinoma from 12 tertiary hospitals across China for external validation. This cohort exhibits a male predominance (5,420/6,668; 81.3\%) and center-level median ages clustered in the late 50s (median of center medians $\approx$\ 59 years; range, 52--61 years). 
For survival analyses, we derived a survival-filtered test set by excluding perioperative deaths, missing follow-up, and incomplete staging, yielding 4{,}190 patients with similar baseline characteristics. The multi-center EMRs capture routine hepatology practice, including ECOG performance status (PS), Child--Pugh score, laboratory indices, tumor markers, imaging findings, and biopsy pathology when available. 
The training data comprise structured tables and clinical narratives for downstream model training (Fig.~\ref{fig:overview}a). Across the 140 synthetic EMRs, three senior hepatobiliary clinicians completed blinded fidelity assessment across six pre-specified dimensions, with NA allowed for non-applicable fields. Across 2,518 non-NA dimension-level ratings, the overall fidelity score was 4.81\,$\pm$\,0.55, with a median of 5 (IQR, 5--5). Clinically acceptable ratings (score $\geq$4) accounted for 96.0\% of all ratings, whereas low-fidelity ratings (score $\leq$2) were uncommon (1.2\%). At the record level, 137/140 synthetic EMRs (97.9\%) achieved a mean score of at least 4. Inter-rater consistency was high, with pairwise within-one-point agreement of 97.1\%, three-rater within-one-point agreement of 95.2\%, and ICC(2,k) of 0.73. These findings support the clinical plausibility and internal consistency of the prompt-augmented EMR narratives (Extended Data Table~\ref{tab:synthetic_emr_fidelity}).

Building on these data, we then trained three families of LLMs (i.e., Qwen3~\cite{qwen3_2025}, QWQ~\cite{qwq32b}, and DeepSeek-R1~\cite{deepseek_r1_2025}) under a two-stage, knowledge-aligned reasoning paradigm (Fig.~\ref{fig:overview}b). \emph{Stage 1: Clinical-Knowledge Familiarization Fine-Tuning:} the model is prompted to act as a hepatobiliary clinician and learn from worked clinical examples (EMR-style narratives infused with guideline logic and evidence), aligning outputs with guideline-consistent reasoning and clinically grounded text. \emph{Stage 2: Experience-Accrual Reinforcement Learning:} reasoning capacity and decision quality are further refined using reinforcement learning with a clinically tailored, verifiable composite reward that jointly optimizes correctness, output-format validity, treatment-ranking consistency, survival estimation, brevity, and CoT quality. The system produces explicit intermediate tags (for example, PS and Child–Pugh score) and returns a risk score-based staging, a ranked list of recommended therapies with evidence-based rationale, and patient-specific survival estimates (Fig.~\ref{fig:overview}c).

\subsection{Evaluation of treatment recommendation performance}
\begin{figure}[t]
    \centering
    \includegraphics[width=0.96\linewidth,height=0.68\textheight,keepaspectratio]{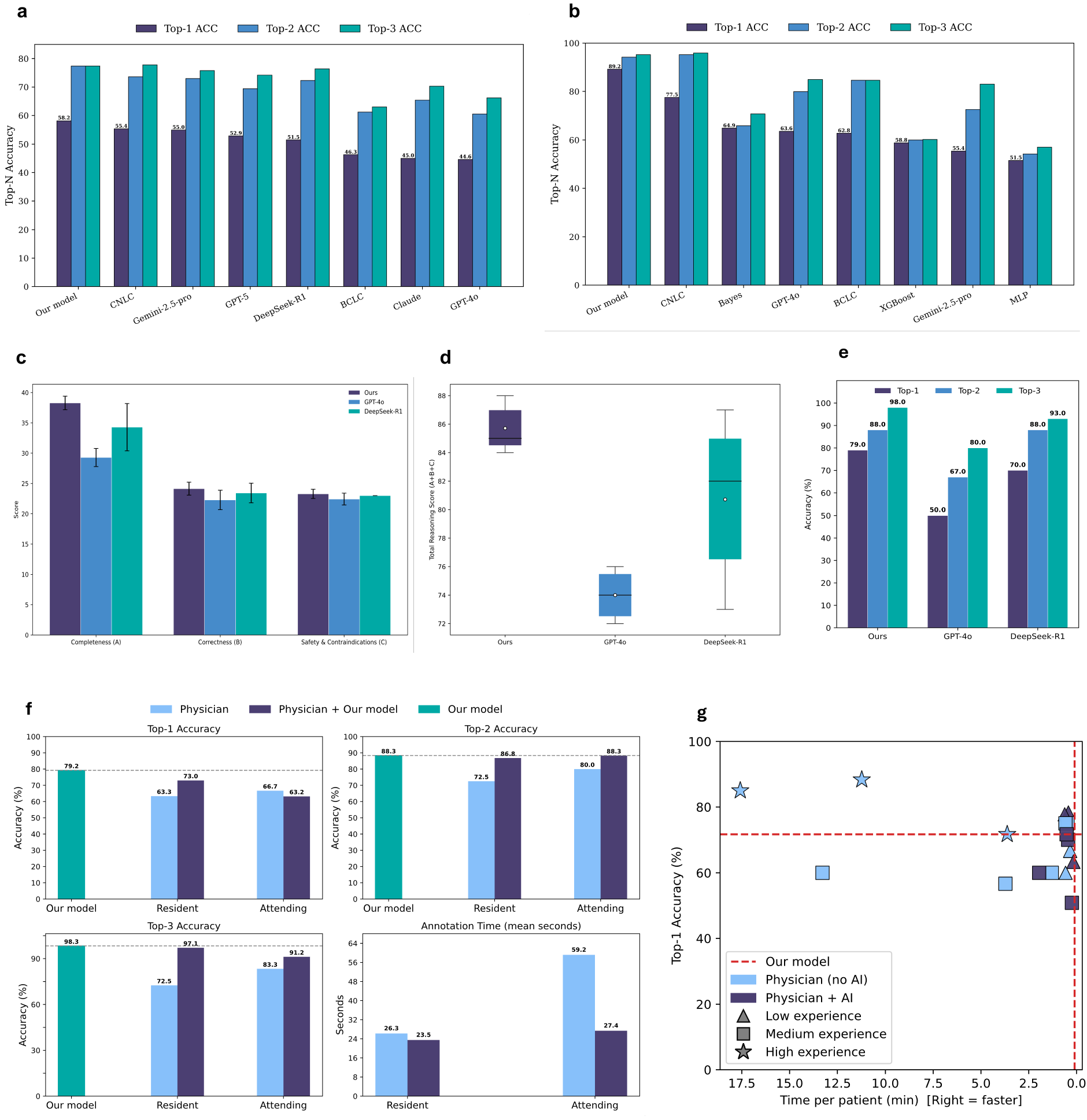}
    \caption{The performance comparison of treatment recommendations among models and physicians. \textbf{a, b}, Top-$k$ accuracy: our model vs. clinical staging systems, ML methods and LLMs. Each model returns a ranked list of candidate therapies, and accuracy-{$k$} counts a hit when any ground-truth appears within the Top-$k$ with $k\in\{1,2,3\}$. \textbf{c, d}, Blinded expert scoring of clinical reasoning and recommendations for our model and other LLMs. c: Three senior hepatobiliary specialists rated CoT and treatment recommendations for hepatocellular carcinoma across three domains (i.e., completeness, correctness, and safety) using a consensus questionnaire aligned with routine practice. {d}: Composite quality score (QNS; sum of the three domains) for our model and other LLMs. \textbf{e}, Accuracy comparison of representative LLMs on the same cases. Blinded expert scoring of clinical reasoning and recommendations. \textbf{f}, Accuracy comparison of our model and human physicians stratified by experience on 60 blinded cases. \textbf{g}, Comparison of Top-1 accuracy and decision time among physicians with or without model assistance on 60 blinded cases.}
    \label{fig:ai_vs_physicians}
\end{figure}

We first assessed the quality of treatment recommendations using a rank-aware Top-$k$ accuracy that measures both list overlap and rank concordance. Across the external multi-center cohort and the internal SEER test set, Fig.~\ref{fig:ai_vs_physicians}a and Fig.~\ref{fig:ai_vs_physicians}b show that HCC-STAR achieves the highest Top-$k$ accuracy for $k\in{1,2,3}$ compared with clinical guidelines (BCLC, CNLC) and representative LLMs (GPT-5, Gemini-2.5-pro, GPT-4o, Claude). Especially, our model significantly outperforms GPT-5 and Gemini-2.5-pro, which verifies the effectiveness of reinforcement learning on HCC-specific clinical knowledge and case-based clinical reasoning. 
On the internal SEER set, it also outperforms traditional machine-learning baselines (e.g., support vector machines, XGBoost) and other methods. Gains are most pronounced for Top-1 accuracy, which is clinically meaningful because the first-line option often determines the treatment plan and downstream prognosis. The rank-aware similarity likewise indicates greater overlap and closer rank agreement with the guideline-consistent reference lists. 

We then examined clinical reliability through two physician-centric evaluations on 60 carefully curated HCC cases, and patient statistics (age, staging, ECOG PS, Child–Pugh) are summarized in Extended Data Fig.~\ref{fig:60_cases_stats}. In a blinded expert scoring study, three senior hepatobiliary specialists independently reviewed and scored anonymized chains of thought and recommendations from our model and two leading LLMs. {HCC-STAR} achieved the highest composite quality score and the highest \emph{safety-pass} rate with the lowest proportion of red-flagged options (Fig.~\ref{fig:ai_vs_physicians}c,~\ref{fig:ai_vs_physicians}d) over DeepSeek-R1 and GPT-4o. The largest margin was in \emph{completeness}, assessed by an eight-item coverage checklist requiring explicit documentation of the staging reference (BCLC or CNLC), ECOG performance status, Child–Pugh score, tumor burden, vascular invasion (including PVTT grade), extrahepatic or nodal status, portal hypertension or FLR adequacy, and transplant eligibility (Supplementary Table~\ref{tab:completeness}). The disproportionate margin on completeness mirrors the EARL training signal: because the composite reward verifies intermediate clinical tags (performance status, Child–Pugh grade, tumor burden, vascular invasion, metastasis, and BCLC/CNLC/TNM stage) rather than only the final recommendation, the model is incentivized to surface exactly the determinants that human raters subsequently check for during clinical review.
Higher CoT scores reflect more complete, guideline-aligned reasoning, which improves auditability of the decision path and reduces low-value detours. In practice, higher correctness and safety scores translate into fewer unsafe options for complex profiles (for example, main-trunk PVTT, significant portal hypertension, or inadequate FLR) and clearer justification when prioritizing resection or ablation, locoregional therapy, or systemic therapy as appropriate. Together, these properties support safer and more transparent recommendations, facilitating shared decision-making at the point of care.

To further evaluate the reliability of the model’s evidence-based justifications, three senior hepatobiliary specialists independently scored the supporting evidence for each recommendation along three dimensions (Supplementary Section~\ref{sec:score_evidence} details the scoring rubric): \emph{completeness} (coverage of required sources, including relevant guideline sections and evidence levels), \emph{correctness} (agreement between the claim and the cited source), and \emph{consistency} (appropriateness for the patient’s clinical context). {HCC-STAR} achieved the highest scores across all three dimensions in 60 cases, outperforming DeepSeek-R1 and GPT-4o (Extended Data Fig.~\ref{fig:evidence_scores}). Together with the CoT evaluation, these findings indicate that the model provides more complete, precise, and reproducible evidence trails, supporting safer and more reliable treatment decision-making. We also compared treatment recommendation accuracy on the same 60 cases (Fig.~\ref{fig:ai_vs_physicians}e), and {HCC-STAR} consistently outperformed both models across ranking metrics, achieving the highest Top-1 (79\%), Top-2 (88\%), and Top-3 (98\%) accuracy. These results demonstrate superior identification of appropriate therapies and closer agreement with expert rankings, indicating that the model delivers clinically reliable reasoning alongside more guideline-concordant treatment prioritization, helping bridge the gap between LLM reasoning and expert-level decision-making. The survival-discrimination ablation (Extended Data Fig.~\ref{fig:cindex_sft_rl}) provides direct evidence that the EARL stage contributes substantively beyond SFT, with the largest gains observed on the external multi-center cohort—consistent with the view that the verifiable, ranking-aware composite reward encourages transferable reasoning rather than memorization.

\subsection{Comparison with physicians in treatment decision-making}
In a separate comparison of LLM and physicians stratified by experience on 60 curated cases, Fig.~\ref{fig:ai_vs_physicians}f illustrates the mean accuracy of participating residents and attending physicians. {HCC-STAR} achieved accuracy closest to that of senior specialists and delivered higher Top-1 accuracy than both junior and intermediate physicians \textit{79.2\% vs. junior (resident) 63.3\%; vs. intermediate (attending) 66.7\%}, and it outperformed other LLMs (best LLM, 70.0\%; Fig.~\ref{fig:ai_vs_physicians}e). Advantages persisted at Top-2 (\textit{ 88.3\% vs. junior (resident) 72.5\%; vs. intermediate (attending) 80.0\%}) and Top-3 (\textit{ 98.3\% vs. junior (resident) 72.5\%; vs. intermediate (attending) 83.3\%}). Especially, improvements in Top-1 accuracy were significant because the first recommended therapy often guides the actual treatment plan and influences prognosis. 
These gains reflect the model’s consistent application of critical decision rules, including identifying candidacy for resection or ablation when eligibility criteria were met, triggering transplant referral when appropriate, and suppressing options that conflicted with absolute or practical contraindications (for example, avoiding inappropriate TACE in main-trunk PVTT). By standardizing rule application across cases, the model can reduce under-treatment in eligible candidates and over-treatment in ineligible patients, thereby improving both sensitivity and specificity across physician strata.

\subsection{Evaluation of LLM-assisted treatment decision-making}
We further assessed our system as a clinical assistant by comparing physicians’ decisions with and without model support and reported the corresponding results in Fig.~\ref{fig:ai_vs_physicians}f and Fig.~\ref{fig:ai_vs_physicians}g. When provided with our model’s chain-of-thought reasoning and treatment recommendations, physicians at almost all levels demonstrated significantly improved accuracy. Specifically, junior (resident) improved Top-1 accuracy from 63.3\% to {73.0\%} ($\Delta_{\text{Top-1}}=+9.7$). For intermediate (attending), Top-1 accuracy was similar with and without assistance ({66.7\%} vs. {63.2\%}). Gains are prominent at Top-2 and Top-3 accuracy. At Top-2, residents improved accuracy from {72.5\%} to {86.8\%} ($\Delta_{\text{Top-2}}=+14.3$) and attendings improved from {80.0\%} to {88.3\%} ($\Delta_{\text{Top-2}}=+8.3$). At Top-3, residents’ accuracy increased by {24.6\%} and attendings’ accuracy increased by {7.9\%}. Importantly, with our model’s assistance, junior physicians achieved decision accuracy that matched or exceeded that of more-experienced colleagues on the same cases (e.g., Resident with AI (73.0\% Top-1 accuracy)  vs. Attending baseline (66.7\%)). In addition to accuracy gains, decision efficiency also improves. Concretely, the average time to reach a treatment decision is reduced: decision time decreased from {26.3\,s} to {23.5\,s} for residents ($\Delta t=-2.8$\,s) and from {59.2\,s} to {27.4\,s} for attendings ($\Delta t=-31.8$\,s). These findings demonstrate the dual benefits of our model as a clinical assistant, enhancing decision quality while reducing workload, particularly for less-experienced physicians.

At the per-physician level (Fig.~\ref{fig:ai_vs_physicians}g), the pre–post Top-1 scatter shows a clear above-diagonal pattern: most physicians shift upward (higher Top-1 with assistance) and leftward (shorter decision time). Improvements are evident across both residents and attendings, with the largest gains among physicians who started at a lower baseline and smaller but consistent gains among high performers.
For instance, as illustrated in Fig.~\ref{fig:case_single_en}, the physician initially favored ablation for a 44-year-old man (ECOG~PS~0, Child--Pugh~A) with a solitary 1.2\(\times\)2.0\,cm subcapsular S3 HCC. With HCC-STAR assistance, the model’s CoT surfaced guideline-concordant staging (CNLC~Ia / BCLC~A / TNM~T1a), emphasized the technical resectability of segment~III, and attached evidence strength to each option (Surgical resection: {Level~1, Recommendation~A}). The physician consequently revised the plan to \textit{surgical resection} (the ground-truth first-line), yielding a Top-1 correction and a marked reduction in decision time (from 78.56\,s\ to 18.05\,s). These results suggest that physicians can benefit from the model’s highlighting of decisive cues (e.g., resectability, transplant deprioritization, perioperative antiviral considerations), thereby improving both decision quality and efficiency. Assistance also reduces between-physician variability and eliminates any accuracy–time trade-off, yielding higher accuracy with shorter decision times.

\subsection{Evaluation of prognostic performance and risk stratification}
\FloatBarrier

\begin{figure}[t]
    \centering
    \includegraphics[width=0.98\linewidth]{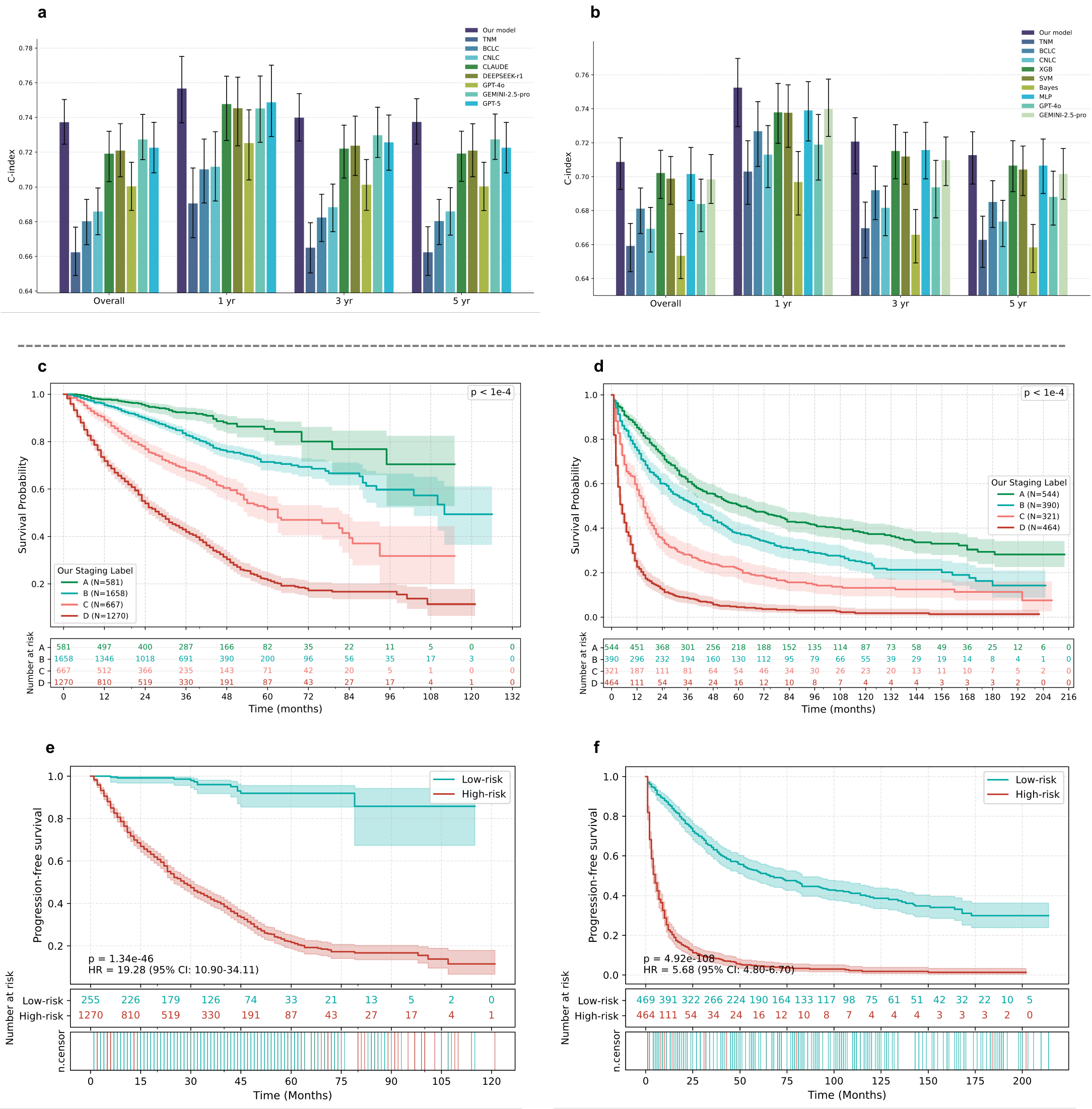}
    \caption{Harrell’s concordance index (C-index) and Kaplan–Meier analysis on the external and internal test cohorts. 
    \textbf{a, b,} Time-dependent Harrell’s C-index for overall survival on the external multi-center cohort and the internal (SEER) cohort at 1, 3, and 5 years. Bars denote means and error bars show 95\% bootstrap confidence intervals; two-sided paired $P$ values for comparisons against our model are reported in Extended Data Table~\ref{tab:cindex_auc_external} and \ref{tab:cindex_auc_internal}. Baselines include clinical staging systems (AJCC/TNM, BCLC, CNLC), representative LLMs, and classical ML methods. 
    \textbf{c, d,} Kaplan–Meier curves on the multi-center cohort and the internal cohort by \emph{model-predicted staging} (A–D) derived from quantile thresholds of the model’s continuous risk score ($K{=}3$ learned on the training set). Curves display a monotonic ordering (A$>$B$>$C$>$D) with early and persistent separation; shaded bands indicate 95\% confidence intervals, log-rank $P{<}10^{-4}$. Hazard ratios and $P$ values are annotated on the plots, and the numbers at risk are reported below each panel. 
    \textbf{e, f,} Kaplan–Meier analysis of progression-free survival on the multi-center cohort and the internal cohort using a pre-specified low- vs.\ high-risk cutoff from the training distribution. A single threshold generalizes across cohorts and yields wide, stable gaps between risk groups; shaded areas show 95\% confidence intervals, two-sided log-rank $P$ values and Hazard ratios are reported.}
    \label{fig:cindex_survival_risk}
\end{figure}

\textbf{Prognostic discrimination by C-index.}
On the external multi-center cohort (Fig.~\ref{fig:cindex_survival_risk}a), HCC-STAR achieved a higher overall Harrell’s concordance index (C-index) (\textit{0.7371; 95\%~CI: 0.7234–0.7507}) than each guideline-based staging system, which clustered around 0.66-0.68, and outperformed a panel of strong general-purpose LLMs (0.70-0.73, including Claude, DeepSeek-R1, Gemini-2.5-pro, GPT-5, and GPT-4o). Two-sided paired $P$ values confirm the statistically significant gains of our model over three guideline-based staging systems (Extended Data Table~\ref{tab:cindex_auc_external}). Similar advantages were observed at 1-, 3-, and 5-year horizons, where HCC-STAR consistently maintained the highest C-index among all methods. On the internal SEER test set (Fig.~\ref{fig:cindex_survival_risk}b), our model likewise achieved superior overall discrimination (\textit{0.7079; 95\%~CI: 0.6935–0.7228}) compared with TNM, BCLC, and CNLC (0.66–0.68) and exceeded classical machine-learning baselines trained on the same structured variables (XGBoost, MLP, and SVM). The reported 95\% confidence intervals and $P$ values (Extended Data Table~\ref{tab:cindex_auc_internal}) demonstrate statistically significant gains of our methods. We further compared C-index across supervised fine-tuned backbones (Extended Data Fig.~\ref{fig:cindex_sft_rl}), revealing a clear and consistent gap in favor of the RL-enhanced model, which supports the notion that survival-aware reinforcement learning improves prediction beyond supervised fine-tuning alone. Collectively, these results suggest that HCC-STAR captures patient-level heterogeneity beyond discrete staging and leverages a continuous, survival-oriented risk representation to maintain strong generalization across multi-center cohorts and time horizons.

\textbf{Prognostic discrimination by ROC.}
Time-dependent receiver operating characteristic analyses (Extended Data Fig.~\ref{fig:rocs_all}) further corroborate these findings. Across 1-, 3-, and 5-year horizons, HCC-STAR achieved the highest area under the receiver operating characteristic curve (AUROC) on both the internal SEER test set and the external multi-center cohort, with absolute gains of roughly 0.07–0.1 over TNM, BCLC, and CNLC at each horizon. Improvements are particularly pronounced at longer horizons, where within-stage heterogeneity becomes more influential and fixed guideline thresholds struggle to separate low- and high-risk patients. Corresponding AUROC estimates, 95\% confidence intervals, and $P$ values (Extended Data Table~\ref{tab:cindex_auc_external} and Table~\ref{tab:cindex_auc_internal}) confirm statistically significant gains over clinical staging systems and competitive LLM baselines. Together with the C-index analyses, these results highlight that a continuous, patient-level survival estimate that is learned from HCC-specific data and refined by reinforcement learning under guideline-aligned objectives supports sharper time-specific discrimination than discrete guideline staging alone.

\textbf{Kaplan–Meier analyses of model-predicted stagings and risk groups.}
Using quantile-based thresholds derived from the training data (quantile $K{=}3$, four risk tiers A-D), our model stratified survival in both the internal and external cohorts. The Kaplan–Meier curves exhibit a clear, persistent, monotonic ordering from A to D (Fig.~\ref{fig:cindex_survival_risk}c,~\ref{fig:cindex_survival_risk}d), with early separation that is maintained throughout follow-up. Confidence bands show minimal overlap for adjacent strata across most time horizons, indicating robust separation even as censoring increases later in follow-up. The pattern is consistent across datasets, despite different case-mixes (e.g., larger B/D groups externally and more balanced strata internally), suggesting that a single set of risk thresholds generalizes well and preserves ranking under distribution shifts. Clinically, survival decreases stepwise from stage~A to stage~D: stage~A maintains the highest survival with a gradual decline; stage~B follows an intermediate course with stable separation from A; and stages~C/D exhibit markedly higher early event rates with sustained gaps thereafter. The absence of curve crossings and the strong log-rank signal ($p{<}10^{-4}$, annotated in each panel) are consistent with approximately proportional hazards across strata.
 


We further stratified patients into low- and high-risk groups using prespecified cutoffs (the 25th and 95th percentiles) derived from the training data and then applied the same thresholds to the internal SEER test set and the external multi-center cohort. The Kaplan–Meier curves display clear divergence between risk groups in both cohorts, with narrow confidence bands for the low-risk group and substantially higher event rates for the high-risk group (Fig.~\ref{fig:cindex_survival_risk}e,~\ref{fig:cindex_survival_risk}f). At 2 years, progression-free survival (PFS) is \textbf{74.0\%} (internal) and \textbf{99.2\%} (external) in the low-risk group, versus \textbf{12.4\%} and \textbf{54.0\%} in the high-risk group, respectively. Hazard ratios for the external and internal cohorts are 19.28 (95\% CI: 10.90-34.11; $P =1.34 \times 10^{-46}$) and 5.68 (95\% CI: 4.80-6.70; $P = 4.92 \times 10^{-108}$) between the low- and high-risk groups, indicating a markedly higher event risk in the high-risk group and providing strong evidence of robust prognostic separation across cohorts. Together, these results support the use of a single decision threshold for cohort-agnostic risk stratification and suggest practical clinical utility for triage, tailoring surveillance intensity, and prioritizing treatment escalation or trial referral in patients flagged as high risk.

\FloatBarrier

\begin{figure}[t]
    \centering
    \includegraphics[width=0.98\linewidth]{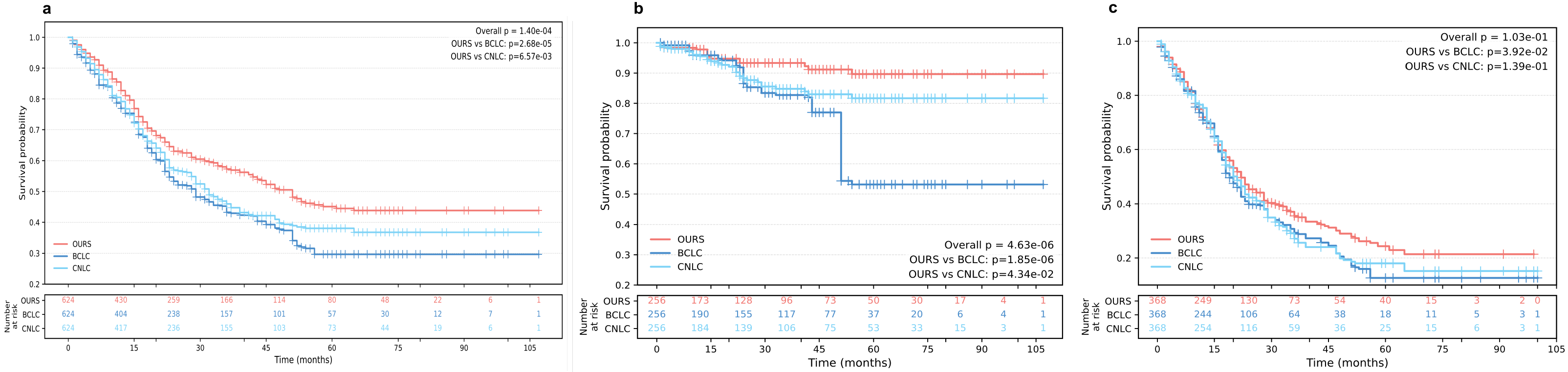}
    \caption{Hypothetical Kaplan–Meier estimated overall survival curves (OS) under alternative treatment recommendations: our model vs. BCLC/CNLC scheme based on the external testing cohorts. \textbf{a}, External cohort overall: counterfactual OS if patients were treated per our model, BCLC, or CNLC. \textbf{b}, Low-risk subgroup defined by our model. \textbf{c}, High-risk subgroup defined by our model. For patients not treated as recommended, outcomes were imputed by random sampling from the external test set. Two-sided log-rank $P$ values compare our model with BCLC/CNLC and numbers at risk are shown.}
    \label{fig:survival_os}
\end{figure}

\textbf{Hypothetical OS analyses under alternative treatment recommendations.}
To compare the prognostic effects of treatment recommendations at the cohort level, we followed prior work~\cite{yau2014development} and constructed hypothetical overall survival (OS) curves for our model and for BCLC and CNLC on the external cohort (Fig.~\ref{fig:survival_os}a). For patients who did not receive the recommended therapy, we imputed outcomes by random sampling from patients in the test set who did receive that therapy. Under this counterfactual assignment, the median OS for our model is \textbf{51} months, in contrast to \textbf{29} and \textbf{32} months when patients receive treatment according to BCLC and CNLC, respectively. The Kaplan–Meier estimates indicate that adherence to our model’s recommendations is associated with longer survival than adherence to BCLC or CNLC. 
Because outcomes are imputed from similar treated cases, these estimates are subject to confounding by indication and should be regarded as hypothesis-generating rather than causal. Whether such patterns would translate into population-level benefits in clinical practice is a question that warrants prospective investigation.

\begin{figure}[t]
\centering
\begin{minipage}{0.99\linewidth}
  \begin{bandbox}{CaseHeader}
  \textbf{HCC case. Early-stage, subcapsular S3 HCC — Surgical Resection}

  \textit{Summary}: 
  Male, 44; ECOG PS 0; Child--Pugh A; no extrahepatic disease; no PVTT. 
  MRI shows a solitary subcapsular lesion in segment~3, \(\sim\)1.2\(\times\)2.0\,cm, arterial hyperenhancement with portal/venous washout and hepatobiliary-phase hypointensity—radiologically consistent with HCC on a cirrhotic background. No radiologic signs of portal hypertension or biliary dilation. 
  Background HBV infection (HBsAg+); Labs: AFP: 31.17\,ng/mL\(\uparrow\); PT: 24.9\,s\(\uparrow\) with INR 0.99; ALT: 37.9\,U/L. 
  
  Reference staging: CNLC~Ia; BCLC~A; TNM~T1a (single \(\le\)2\,cm, no vascular invasion or metastasis).
  \end{bandbox}

  \vspace{2pt}
  \begin{bandbox}{ClinInit}
  \textbf{Physician (Attending without our model)}: Favors \textbf{ablation} as first-line over resection; transplant not considered.
  \end{bandbox}

  \vspace{2pt}
  \begin{bandbox}{ModelCoT}
  \textbf{Model’s chain of thought (excerpt)}:

  \textit{Staging check}. CNLC: PS~0, CP-A, single lesion \(\le 3\,\mathrm{cm}\), no PVTT or metastasis \(\Rightarrow\) \textbf{CNLC~Ia}. 
  BCLC: single small tumor, CP-A, PS~0 \(\Rightarrow\) \textbf{BCLC~A}. 
  TNM: single lesion \(\le 2\,\mathrm{cm}\), no vascular invasion \(\Rightarrow\) \textbf{T1a}. 
  Subcapsular S3 location is technically resectable (wedge/segment~III) with adequate functional reserve.

  \medskip
  \textit{Candidate options under CNLC~Ia} (\( \{\)resection, ablation, transplant\(\}\)):

  \textbf{1) Surgical resection}. 
  Indicated for single small HCC with CP-A, PS~0, and resectable anatomy; curative-intent first-line in CNLC~Ia. 
  For lesions \(\le 3\,\mathrm{cm}\), overall survival is comparable to or slightly better than ablation; resection reduces local recurrence when margins are secure and FLR is adequate. 
  \emph{Evidence}: \textbf{Level~1, Recommendation~A} (CNLC early-stage algorithm; resection preferred when operable). 

  \textbf{2) Ablation}. 
  Appropriate for single lesions \(\le 3\,\mathrm{cm}\) when safely accessible; subcapsular S3 is feasible with image guidance and tract cauterization, but may carry relatively higher risk of local recurrence than an R0 resection. 
  Ablation is a reasonable alternative if surgical risk increases or margins are challenging. 
  \emph{Evidence}: \textbf{Level~1, Recommendation~A} (CNLC early-stage local therapy). 

  \textbf{3) Liver transplant}. 
  Reserved for decompensated function or patients not amenable to resection and ablation; organ scarcity prioritizes more advanced hepatic dysfunction. In this case (CP-A, single small tumor), transplant is not prioritized. 
  \emph{Evidence}: \textbf{Level~2, Recommendation~A} (indicated mainly for poor hepatic reserve or downstaging pathways).

  \medskip
  \textit{Safety/feasibility checks}. 
  Verify absence of significant portal hypertension (platelets, splenomegaly, varices), confirm adequate FLR and vascular/biliary proximity. 
  HBV antiviral therapy perioperatively is advised to reduce reactivation risk. 
  Coagulation indices acceptable given INR~0.99 (prolonged PT with normal INR likely methodological; re-check before OR). 

  \textbf{Recommendation}: \textbf{Resection} \(>\) \textbf{Ablation} \(>\) \(\,\)Transplant (not prioritized).
  \end{bandbox}

  \vspace{2pt}
  \begin{bandbox}{Revised}
  \textbf{Physician (Attending with our model)}: Agree with the model’s reasoning process and rationale; \textbf{surgical resection} is preferred; \textbf{ablation} as an alternative if surgical constraints arise. 
  \textit{Accuracy} \& \textit{efficiency}: Top-1: \(0 \rightarrow 1\). Decision time: \(78.56\)\,s \(\rightarrow\) \(18.05\)\,s.
  \end{bandbox}
\end{minipage}
\caption{\textbf{Representative case with model assistance.} 
Orange band: case header and key facts; green: physician decision without our model; grey: model's chain of thought with per-option evidence/strength; blue: revised physician decision with our model plus changes in Top-1 accuracy and decision time.}
\label{fig:case_single_en}
\end{figure}

Stratification by model-defined risk produced consistent patterns (Fig.~\ref{fig:survival_os}b,~\ref{fig:survival_os}c). In the low-risk group, hypothetical OS under our model’s recommendations remained the highest throughout follow-up, with early curve separation that widened over time, and log-rank tests versus BCLC and CNLC are statistically significant. In the high-risk group, overall survival was shorter across all strategies. Yet, our model still yielded a right-shifted curve with a longer tail, and the differences are smaller, with some intervals only being borderline significant. Collectively, these subgroup analyses indicate that the potential benefit associated with adhering to our model’s recommendations is maintained across risk strata and may be most pronounced among patients with more favorable baseline profiles.

\subsection{Evaluation of adaptability to guideline updates}
\begin{figure}[t]
    \centering
    \includegraphics[width=0.99\linewidth]{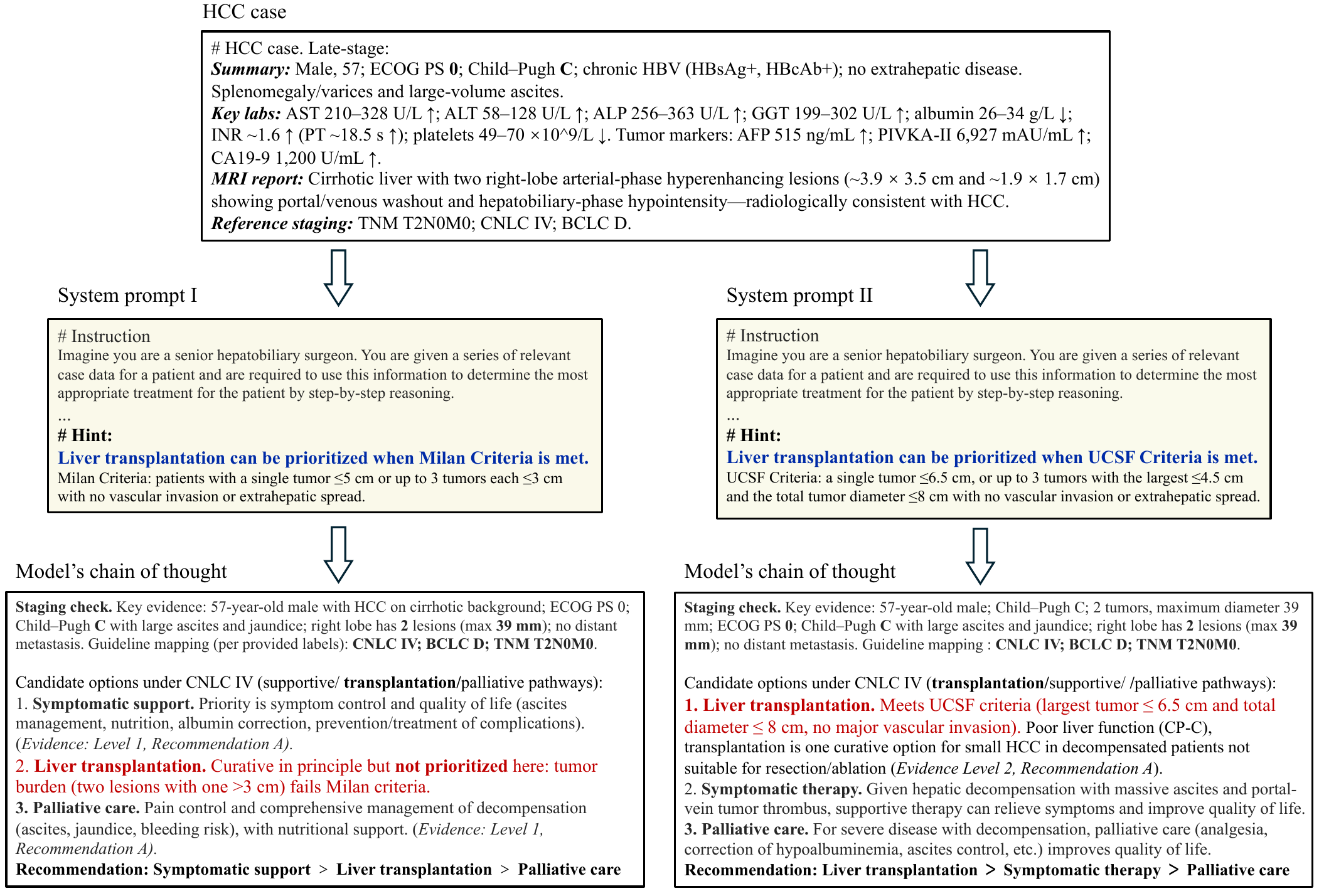}
    \caption{Prompt-conditioned generalization of treatment recommendations without retraining. We show (top) a representative HCC case; (middle) two alternative system prompts encoding liver transplantation standards (Milan vs. UCSF criteria); and (bottom) the model's chain of thought with candidate options and prioritization under CNLC IV. Switching from Milan to UCSF via a lightweight prompt edit reorders the recommendation (transplantation de-prioritized vs. prioritized) and updates the accompanying rationale while all other case inputs remain fixed. \textit{This liver-transplant example is presented solely to illustrate the model's generalization via system-prompt adjustments and does not engage in, or substitute for, medical or ethical decision-making.}}
    \label{fig:trans}
\end{figure}

Prior work identifies two mechanisms by which models with strong reasoning generalize to new scenarios: verification-aware reinforcement learning that enables cross-domain transfer to out-of-distribution problems~\cite{xie2025logicrl}, and instantiated training data that provides contextual grounding for robust generalization~\cite{hua2025disentangling}. These observations are consistent with the emerging SFT–RL paradigm, where recent studies reveal a critical distinction: supervised fine-tuning primarily encodes factual memorization from training data, while reinforcement learning cultivates transferable reasoning patterns~\cite{kang2024learning}. Following this paradigm, we precondition the model with instantiated clinical scenarios to consolidate guideline knowledge via SFT, then leverage reinforcement learning over the same scenario space to develop adaptive reasoning capabilities. This strategy is particularly valuable when clinical guidelines undergo incremental updates in treatment protocols or procedural specifications. Rather than requiring full retraining, the reasoning capabilities in our HCC-STAR could enable the model to adapt to these guideline changes through targeted system prompt adjustments, effectively achieving knowledge transfer and generalization. We illustrated this capability with a liver transplant case (Fig.~\ref{fig:trans}), where applying different liver transplantation standards (Milan versus UCSF criteria) required only lightweight system prompt refinement to update the recommended pathway and its rationale without retraining the model. While this example serves as a proof of concept, a broader, quantitative evaluation of update fidelity across diverse guideline-change scenarios would further substantiate prompt-based adaptability and is a natural direction for future work.

\section{Discussion}\label{sec3}
This study presents \textit{HCC-STAR}, a domain-adapted \emph{reasoning} model that unifies fine-grained HCC staging, evidence-based precision treatment recommendation, and individualized survival prediction within a single, clinically aligned framework. We synthesized EMR-style narratives from SEER using a prompt-based data augmentation strategy informed by distilled clinical knowledge from clinical guidelines and real-world EMRs. We adopted a \emph{knowledge-aligned reasoning} paradigm with two stages: \emph{Clinical-Knowledge Familiarization Fine-Tuning (CKF-FT)} and \emph{Experience-Accrual Reinforcement Learning (EARL)} with GRPO~\cite{shao_deepseekmath_2024} under a clinically \emph{verifiable} composite reward. The system surfaces transparent intermediate tags and patient-level outputs and consistently outperforms major staging systems and representative LLMs across an internal SEER test set and a large external cohort from 12 tertiary hospitals across China, with consistent gains in the clinical evaluation and blinded scoring.

First, the model demonstrates superior discrimination for overall survival, achieving higher time-dependent Harrell’s concordance index (C-index) and AUROC than AJCC/TNM, BCLC, and CNLC in both internal and external cohorts, with advantages persisting at the 1-, 3-, and 5-year endpoints. Kaplan–Meier analyses based on model-predicted stagings and prespecified low- and high-risk groups show clear and monotonic separation. These findings indicate that continuous, patient-level risk estimation preserves within-stage heterogeneity that rule-based staging compresses into coarse categories.
Second, for treatment recommendations, the system attains the highest Top-$k$ accuracy against guideline systems, representative LLMs, and classical ML baselines. Gains are most pronounced for Top-1, which is clinically meaningful because the first-line recommendation typically determines the actual plan and downstream prognosis. In blinded expert scoring, the quality of CoT improves most in \emph{completeness} and is accompanied by higher safety-pass rates and fewer red-flag options with more reliable evidence-based justification. In model-physician comparisons, the model achieves accuracy close to senior specialists, surpassing junior and intermediate physicians, and further enhances the accuracy and efficiency of less experienced clinicians when used as an assistant.

Methodologically, two elements are central. First, \emph{domain-adapted data synthesis}: after harmonizing codes and units, clinician-validated prompts expand structured entries from SEER into EMR-style narratives with embedded guideline logic and brief evidence notes, yielding dual representations that supported downstream modeling and auditability, and a blinded multi-dimensional fidelity assessment quantifies their clinical plausibility (Extended Data Table~\ref{tab:synthetic_emr_fidelity}). Second, \emph{training with clinical knowledge and verifiable feedback}: CKF-FT teaches guideline-consistent reasoning from worked clinical examples rather than text-level memorization, while EARL optimizes a composite reward that is step-verifiable (i.e., inspecting intermediate clinical tags rather than only the final answer) and decoupled (i.e., separating treatment-ranking and survival-estimation gradient updates to prevent multi-objective interference). This verifiable, ranking-aware reward design should, in principle, transfer to other oncology tasks where guideline-anchored decision rules and ranked candidate therapies can be formalized. Ablation analyses (Extended Data Fig.~\ref{fig:cindex_sft_rl}) confirm consistent EARL gains beyond supervised fine-tuning, with the largest improvements on the external multi-center cohort.

The system outputs risk score-based staging, ranked therapies with evidence-based rationales, and survival estimation, which can be designed to slot into multidisciplinary team (MDT) discussions and routine hepatology practice. By surfacing eligibility for resection, ablation, or transplant when criteria are met and by suppressing options that conflicted with absolute or practical contraindications such as main-trunk PVTT, marked portal hypertension, or inadequate FLR, the model can reduce both under-treatment and over-treatment. In resource-constrained settings with limited specialist access, decision support may help shorten time-to-treatment and improve individualized triage, addressing real-world delays that are associated with poorer outcomes.

Recent imaging- and language-based systems have advanced AI-driven oncology by improving risk stratification and decision support~\cite{xiang_visionlanguage_2025,HEROVision,Menglei}. Our work extends this line of work in three ways: targeting HCC across the full continuum of staging, treatment, and survival; training on EMR-style narratives synthesized from discrete data with guideline-aligned knowledge injection; and coupling CKF-FT with EARL under clinically verifiable rewards to enhance reasoning quality and transportability. As the guideline logic is exercised during reasoning rather than memorized, incremental guideline updates can be incorporated via system-prompt edits without model retraining. Unlike classical ML models tied to fixed schemas and therefore restricted to the internal set, HCC-STAR ingests unstructured narratives, preserves patient-level context, and shows stronger generalization to external sites without retraining.

This study is retrospective. Despite validation across 12 centers and blinded reader studies, prospective trials are necessary to quantify the real-world impact on clinical management and outcomes, and we will pursue prospective testing in future work. The hypothetical overall-survival analysis relies on outcome imputation from clinically similar treated cases and is vulnerable to confounding by indication, as real-world deviations from guideline-recommended therapy often reflect unrecorded clinical factors (e.g., marginal hepatic reserve, inadequate FLR, severe portal hypertension). These counterfactual estimates should therefore be interpreted as illustrative rather than causal. Documentation practices differed across centers, and part of the EMR-style narratives were synthesized using prompt-based data augmentation, which may introduce the distributional shift relative to native EMRs. Fairness across subgroups, including HBV-predominant versus non-HBV etiologies and marginal hepatic reserve, warrants further evaluation. General-purpose LLMs are released rapidly, and an exhaustive comparison against every new system is infeasible. Our baselines (GPT-5, Gemini-2.5-Pro, GPT-4o, Claude, DeepSeek-R1) were contemporaneous with our Qwen3-series base model to reflect a common technological horizon, and the proposed knowledge-aligned reasoning paradigm is expected to transfer to stronger backbones as they become available. As guidelines evolve, continual learning or parameter-efficient updates will be necessary to maintain alignment while preserving privacy.

The model’s structured outputs and CoT provided guideline-consistent and evidence-based rationales, which may increase clinician trust and facilitate multidisciplinary consensus. Safety monitoring should include guardrails for contraindication checks~\cite{hager2024evaluation}, uncertainty cues\cite{banerji2023clinical} for borderline cases, and institution-level audit trails to ensure transparency and accountability. To measure real-world impact, randomized or pragmatic trials are necessary to assess the effects on treatment selection, time to treatment, survival, quality of life, and cost-effectiveness. Studies of human–AI teaming should determine when assistance most benefits clinicians and how to mitigate over- or under-reliance.

In parallel, extending the framework with imaging and pathology via vision–language models may further improve staging fidelity, vascular invasion assessment, and transplant candidacy estimation. Besides, incorporating longitudinal laboratory results and treatment trajectories could also strengthen dynamic risk prediction. Future work should evaluate subgroup fairness and develop continual-learning strategies~\cite{wang2025hide} as guidelines evolve, including parameter-efficient learning~\cite{ji2024towards} and privacy-preserving alignment~\cite{zhangstair} that align with local documentation styles without centralizing data.

In summary, {HCC-STAR} delivers patient-level, guideline-consistent staging, treatment prioritization, and survival prediction with transparent reasoning and strong external generalization. By unifying tasks that are typically siloed and by adopting a \emph{knowledge-aligned reasoning} paradigm with clinical-knowledge familiarization and accumulation learning under a clinically \emph{verifiable composite reward}, the system advances AI-assisted hepatology and lays the groundwork for prospective deployment and next-generation clinical decision support.

\section{Methods}\label{sec4}
\subsection{Ethical approval}
This study was conducted in accordance with the Declaration of Helsinki~\cite{wma_helsinki_2013}. Approval for the use of de-identified patient data was obtained from the institutional review boards (IRBs) of Beijing Tsinghua Changgung Hospital, Tianjin Medical University Cancer Institute and Hospital, and Xingtai People's Hospital of Hebei Medical University (25532-4-01), Zhongnan Hospital of Wuhan University (2025304K), The Second Affiliated Hospital Zhejiang University School of Medicine (2025-1106), Shenzhen People's Hospital (LL-KY-2025247-01); The Second Affiliated Hospital of Harbin Medical University (KY2025-088), Guangxi Medical University First Affiliated Hospital (2025-K0373), Fourth Affiliated Hospital of Harbin Medical University (2025-ethics-33), and Peking Union Medical College Hospital (JS-1391).

\subsection{Data acquisition and preprocessing}
This study integrates two complementary data sources: (i) American HCC data from the Surveillance, Epidemiology, and End Results (SEER) program~\cite{seer_overview}, and (ii) real-world electronic medical records (EMRs) collected from 12 geographically distinct tertiary hospitals across China.

For SEER, we retrieved hepatocellular carcinoma (HCC) cases diagnosed between 2004 and 2020, together with demographic, pathological, and clinical variables. Demographic features include sex, age at diagnosis, and marital status~\cite{usda_rucc2013}. Tumor-related variables comprise histology (e.g., ICD-O-3: 8170/3, HCC, NOS)~\cite{icdo3_who2013}, tumor grade, TNM staging~\cite{ajcc8_manual}, tumor size, and the number of malignant tumors per patient. Treatment variables include surgical interventions, lymph node surgery, radiotherapy, and other locoregional therapies. Biomarker status (e.g., AFP) is recorded when available. Survival outcomes are derived from overall survival (OS) and cause-specific survival (CSS), with follow-up information on vital status, survival months, and cause of death. Cases with incomplete survival data or ambiguous treatment documentation are excluded. The final SEER fields used for analysis are summarized in Extended Data Table~\ref{tab:seer_fields_grouped}.

In parallel, de-identified electronic medical records (EMRs) provide fine-grained, patient-level clinical details that reflect routine clinical practice in hepatology. Records include Eastern Cooperative Oncology Group performance status (ECOG PS)~\cite{oken_ecog_1982}, Child--Pugh score~\cite{pugh_childpugh_1973}, and explicit documentation of extrahepatic metastasis. Laboratory examinations include complete blood counts (e.g., WBC 6.2~$\times 10^9$/L, hemoglobin 125~g/L, platelets 190~$\times 10^9$/L), hepatic function tests (bilirubin fractions, AST, ALT, alkaline phosphatase, GGT, albumin, globulin), and tumor markers (e.g., AFP 445~ng/mL), alongside viral serologies. Histopathology is available for the subset undergoing liver biopsy, providing a concise morphologic context. Imaging studies routinely include abdominal CT, contrast-enhanced MRI, and ultrasound to support diagnosis and longitudinal assessment. A structured summary of EMR fields is provided in Extended Data Table~\ref{tab:emr_fields_grouped}.

Taken together, the SEER registry provided large-scale structured data characterizing population-level patterns of HCC. In contrast, multi-center EMRs delivered granular patient-level detail, including laboratory, imaging, and histopathology assessments. This dual-source design enabled the construction of structured tables and natural-language clinical narratives, which were subsequently harmonized into instruction-style training corpora for the development of a large language model (LLM).

\subsection{External multi-center patient population}
We assembled an external cohort of 6{,}668 patients with hepatocellular carcinoma (HCC) from 12 tertiary hospitals across China for model external validation, spanning 12 sites: Center A (Beijing Tsinghua Changgung Hospital), Center B (Tianjin Medical University Cancer Institute and Hospital), Center C (Zhongnan Hospital of Wuhan University), Center D (Mengchao Hepatobiliary Hospital of Fujian Medical University), Center E (Xingtai People's Hospital of Hebei Medical University), Center F (The Second Affiliated Hospital Zhejiang University School of Medicine), Center G (Shenzhen People's Hospital), Center H (Southwest Hospital, Third Military Medical University [Army Medical University]), Center I (The Second Affiliated Hospital of Harbin Medical University), Center J (Guangxi Medical University First Affiliated Hospital), Center K (Fourth Affiliated Hospital of Harbin Medical University), and Center L (Peking Union Medical College Hospital). Center-level demographics show a male predominance (5{,}420/6{,}668; 81.3\%). Age distributions are consistent across centers, with median ages typically in the late 50s (median of center medians $\approx$59 years; range, 52–61 years). Among patients with available data, 65.6\% (3{,}917/5{,}972) have cirrhosis, 19.1\% (1{,}141/5{,}981) have diabetes, and 28.6\% (1{,}713/5{,}989) have hypertension.

For downstream evaluation, we defined a survival-filtered testing set by excluding (i) perioperative deaths, (ii) patients with missing follow-up, and (iii) cases with incomplete staging under BCLC, CNLC, or TNM. The resulting set comprises 4{,}190 patients across the same 12 centers and preserved similar baseline patterns (male 82.5\% [3{,}455/4{,}190]; cirrhosis 70.4\% [2{,}866/4{,}073 available]; diabetes 19.5\% [799/4{,}102 available]; hypertension 29.6\% [1{,}220/4{,}116 available]). Extended Data Fig.~\ref{fig:multi_center_flow} depicts the flow of patient inclusion and exclusion. Center-specific characteristics, including ECOG PS, Child–Pugh, mALBI, tumor burden, vascular invasion, nodal/distant metastasis, and treatment histories (e.g., surgery, RFA, TACE, HAIC, EBRT, SIRT, TKI, ICIs), are summarized in Extended Data Table~\ref{tab:external_multicenter_baseline} and Extended Data Table~\ref{tab:external_multicenter_baseline_filtered}. The source data files are provided in the Supplementary Materials.


\subsection{Prompt-based and clinician-validated data augmentation} 
To link structured variables with narrative clinical contexts, and to leverage large-scale SEER data for LLM training, we implemented a prompt-based data augmentation strategy guided by distilled clinical knowledge (such as treatment guidelines and evidence-based justifications) that is derived from both the Chinese Expert Consensus on Conversion and Perioperative Therapy of Primary Liver Cancer (2024 edition) and clinician-validated EMR exemplars curated at Tsinghua Changgung Hospital. For example, we formalized staging and treatment logic into compact decision trees and embedded them in the augmentation prompts (Supplementary Listing~\ref {lst:hcc_prompt_cnlc}), making the reasoning steps explicit and verifiable to support reliable data generation.
Hepatology specialists reviewed representative EMRs and distilled high-fidelity prompt templates that reflect routine documentation patterns, including symptom sequencing, integration of laboratory and imaging findings, the rationale for staging, and the discussion of treatment considerations. We then programmatically mapped SEER variables into these templates to synthesize instruction-style clinical narratives that preserved ground-truth labels while enriching clinical semantics. Using these prompts, a large language model (e.g., GPT-4o) generated EMR-like discharge summaries, staging interpretations, and treatment notes that incorporate key demographic, tumor, and outcome information.

The harmonized dataset was represented in two complementary formats: (i) structured tabular data from SEER for baseline statistical modeling and (ii) synthetic narrative EMRs suitable for LLM training. Fig.~\ref{fig:overview}a presents the overall pre-processing workflow, including data cleaning, prompt-based augmentation to produce EMR-style narratives, and cross-source harmonization to obtain both structured and natural-language records. The Supplementary provided a synthetic EMR-style template for LLM training (Listing~\ref {lst:json_emrs}) and the data augmentation prompt (Listing~\ref{lst:emr_gen_prompt}).

\subsection{Quantitative fidelity assessment of synthetic EMR narratives}
To address the concern that prompt-augmented EMR narratives may diverge from real clinical documentation, we performed a quantitative, blinded fidelity assessment of the synthetic corpus. From the augmented training corpus, we randomly sampled 140 synthetic EMRs, stratified across CNLC stages to ensure balanced coverage of early, intermediate, and advanced disease. Three senior hepatobiliary clinicians, each with at least 10 years of clinical experience and blinded to model identity and case origin, independently scored each EMR along six pre-specified dimensions using a five-point Likert scale (1 = highly unreasonable; 5 = fully reasonable; NA = not applicable). Raters were required to provide a free-text rationale whenever they assigned a score of 3 or below, preserving qualitative evidence of failure modes alongside the numeric scores.

The six dimensions were designed to interrogate both field-level plausibility and cross-field coherence: (i) plausibility of the assigned ECOG performance status conditional on other EMR fields; (ii) plausibility of the Child–Pugh grade given the laboratory and imaging findings; (iii) consistency between the extrahepatic-metastasis flag and the imaging report; (iv) internal consistency of the laboratory panel and its compatibility with the clinical profile of primary liver cancer; (v) realism of the imaging report relative to routine radiologic documentation; and (vi) global narrative consistency across all fields as a single clinical story. The full scoring interface is provided in Supplementary Fig.\ref{fig:cnlc_demo}.

We summarized fidelity in three ways. First, we reported the mean and median Likert score per dimension, together with the proportion of samples receiving a score of 4 or 5 (regarded as clinically acceptable). Second, we computed inter-rater agreement using Krippendorff's alpha (interval distance for five-point Likert scores) across all raters and dimensions, and additionally reported intraclass correlation coefficients (ICC(2,k), two-way random-effects, average measures) for each dimension. Third, we flagged any EMR in which one or more raters assigned a score of 1 or 2 to any dimension as a candidate low-fidelity case; these cases were re-examined to characterize systematic failure modes (for example, laboratory–imaging incoherence or implausible Child–Pugh assignment). Aggregate results are summarized in Extended Data Table~\ref{tab:synthetic_emr_fidelity}. In brief, the overall fidelity score was 4.81\,$\pm$\,0.55 across 2,518 non-NA dimension-level ratings, and 96.0\% of ratings were clinically acceptable (score $\geq$4). Inter-rater consistency was assessed using exact agreement, within-one-point agreement, Krippendorff's $\alpha$, and ICC(2,k); NA ratings were excluded from the corresponding denominators.

\subsection{Development of LLM-based system}
We developed an LLM-based framework for staging, treatment recommendation, and survival prediction that employs a \emph{knowledge-aligned reasoning mechanism} rather than \emph{text-level memorization of guidelines}. Instead of training on verbatim guideline documents, we constructed \emph{worked clinical examples} via prompt-based augmentation that injects guideline logic and evidence citations into realistic EMR-style narratives. We trained the model to behave as a hepatobiliary clinician and to reason across diagnostic notes, imaging reports, operative summaries, and discharge documentation, thereby internalizing guideline-consistent logic through repeated practice instead of text memorization.
This real-world case-based, reasoning-first design enhanced contextualization and supported robust generalization across external datasets and tasks, just as prior works also demonstrated the capability of reinforcement learning in acquiring generalizable knowledge across rule-variant textual and visual settings~\cite{chu2025sft,xie2025logicrl}. Fig.~\ref{fig:overview} summarizes the data flow as well as model development and validation.
Compared with conventional machine learning methods (e.g., multilayer perceptrons and XGBoost) that rely exclusively on structured features, our framework directly ingests heterogeneous clinical narratives. This end-to-end formulation encourages the model to surface clinically salient factors, chain reasoning steps to guideline-aligned actions, and make intermediate variables explicit for verification and reward shaping during reinforcement learning.

\subsection{Model selection and clinical-knowledge familiarization fine-tuning}
In clinical training, physicians do not begin with high-stakes decision-making but build familiarity with the clinical context and guideline logic by working through representative cases, mastering terminology, staging cues, treatment intent, and coarse prognostic patterns. Motivated by this progression, we introduced a clinical-knowledge familiarization phase that mirrored this process, requiring the model to reason under the same constraints rather than memorize phrasing. We frame this stage as \emph{clinical-knowledge familiarization fine-tuning}: the model pre-reads (i.e., skims and works through) a large volume of EMR-style cases to acquire basic knowledge of hepatocellular carcinoma, including terminology, staging cues, treatment intent, and coarse prognostic patterns, before engaging in fine-grained decision optimization.

We explored three families of large-scale models, including Qwen3-8B/32B~\cite{qwen3_2025}, QWQ-32B~\cite{qwq32b}, and DeepSeek-R1~\cite{deepseek_r1_2025}, which were chosen for their strong Chinese biomedical capabilities and robust reasoning. Each model was adapted via supervised fine-tuning (SFT) on a curated corpus of approximately 20{,}000 instruction–response pairs. The corpus integrates (i) guideline-aligned synthetic reasoning cases, (ii) narratives distilled from real-world EMRs, and (iii) SEER-derived augmented mappings, so that guideline logic and evidence citations are embedded within the realistic clinical context rather than presented as abstract prose. 

Formally, given an input case description $x$ and its target output $y$ (treatment recommendations and prognostic information, including survival months), the supervised fine-tuning objective is the token-level cross-entropy:
\begin{equation}
    \mathcal{L}_{\text{SFT}} = - \sum_{t=1}^{T} \log \pi_{\theta}(y_t \mid x, y_{<t}),
\end{equation}
where $T$ denotes the output length, $x$ is the input record, $y_t$ is the reference token at step $t$, and $\pi_\theta$ is the language model parameterized by $\theta$. This objective encouraged the model to imitate guideline-based reasoning before reinforcement learning, providing a stable initialization that mitigated reward exploitation.

To facilitate reinforcement learning in later stages, we employed \emph{structured prompting} and required the model to generate intermediate tags such as \texttt{<ps>}, \texttt{<child\_pugh>}, and \texttt{<tumor\_size>}. This design enabled automatic reward computation while maintaining transparency for human verification. We also embedded an elaborate example into the prompt to improve generalization across heterogeneous patient cases.
We performed clinical-knowledge familiarization fine-tuning on NVIDIA B200 GPUs for three epochs at a learning rate of 5e-6, using gradient checkpointing~\cite{chen_checkpoint_2016} and mixed-precision training~\cite{micikevicius_mixed_2017} to enhance memory and computational efficiency.

\subsection{Experience-Accrual reinforcement learning for LLMs}
Clinical–Knowledge Familiarization Fine–Tuning (CKF–FT) is analogous to the early phase of clinical training, in which a junior trainee consolidates foundational knowledge using textbook-like yet realistic cases. In the next stage, we sought to mirror residency-style practice, where the learner attends to details, integrates patient-specific features, and justifies each recommendation with evidence. Reinforcement learning has been shown to strengthen reasoning beyond SFT~\cite{luowizardmath,shao_deepseekmath_2024,chen2025bridging,ouyang2022training} by optimizing behavior with task-specific rewards. Therefore, we performed \emph{Experience-Accrual Reinforcement Learning (EARL)} with a multi-dimensional and verifiable reward design to enhance the model's clinical understanding and reasoning abilities. Under EARL, the model received fine-grained case-level feedback, attended to subtle contraindications and comorbidities, generated evidence-based rationales, ranked therapies appropriately rather than collapsing onto a single option, and learned individualized prognostic mappings via outcome-aware feedback. This learning trajectory, from clinical knowledge familiarization to experience accrual, parallels real-world training and reflects the progression from junior to senior physician.



Based on the supervised fine-tuning performance of different LLMs (Extended Data Fig.~\ref{fig:sfts_compare}), we determined Qwen3-32B as the backbone of the reasoning model owing to its superior clinical knowledge alignment performance, and subsequently optimized it under a composite reward.
To enable precision therapy in HCC, we designed a verifiable, multidimensional reward tailored to guideline-consistent treatment recommendations and survival prediction.
In contrast to single-objective rewards that only evaluate correctness, our composite design jointly evaluated \textit{accuracy}, \textit{clinical interpretability}, \textit{guideline consistency}, and \textit{clinical evidence}, encouraging outputs that better reflect real-world clinical reasoning. 
The RL objective maximized the expected reward under the learned policy:
\begin{equation}
\max_{\theta}\ \mathbb{E}_{\,x\sim\mathcal{D},\ \hat y\sim \pi_{\theta}(\cdot\mid x)}\big[\,R(\hat y, y)\,\big],
\label{eq:rl-objective}
\end{equation}
where $R(\hat y,y)$ denotes the composite reward defined as follows.

\begin{enumerate}
\renewcommand{\labelenumi}{(\arabic{enumi})}
    \item \textbf{Process Reward} ($R_{\text{proc}}$).  
    This reward verifies intermediate reasoning by checking whether nine key diagnostic attributes are extracted and labeled correctly: \texttt{<ps>}, \texttt{<child\_pugh>}, \texttt{<metastasis>}, \texttt{<cancer\_thrombus>}, \texttt{<num\_tumor>}, \texttt{<tumor\_size>}, \texttt{<cnlc>}, \texttt{<bclc>}, and \texttt{<tnm>}, which together capture patient status, tumor burden and progression, and clinical staging. 
    Formally, let
     \[
\mathcal{F}_{\mathrm{proc}} =
\left\{
\begin{array}{l}
\texttt{ps},\ \texttt{child\_pugh},\ \texttt{metastasis},\\
\texttt{cancer\_thrombus},\ \texttt{num\_tumor},\ \texttt{tumor\_size},\\
\texttt{cnlc},\ \texttt{bclc},\ \texttt{tnm}
\end{array}
\right\}.
\]
    For each $f_k \in \mathcal{F}_{\mathrm{proc}}$, let $\hat f_k$ denote the model prediction and $f_k$ denote the ground truth. 
    The reward is computed as
    \begin{equation}
        R_{\text{proc}} = \frac{1}{9} \sum_{k=1}^{9} \mathbb{I}\{\hat f_k = f_k\},
    \end{equation}
    where $\mathbb{I}\{\cdot\}$ is the indicator function. 
    This encourages explicit reasoning over clinically relevant variables, improving transparency and downstream interpretability.

    \item \textbf{Format Reward} ($R_{\text{fmt}}$).  
    This component checks the presence, uniqueness, and validity of all required structural tags, thereby improving controllability and reducing hallucinations. 
    The required tag set is
    \[
\mathcal{T}_{\mathrm{req}} =
\left\{
\begin{array}{l}
\texttt{ps},\ \texttt{child\_pugh},\ \texttt{metastasis},\ \texttt{cancer\_thrombus},\\
\texttt{num\_tumor},\ \texttt{tumor\_size},\ \texttt{treatment},\ \texttt{treatment\_not\_recommended},\\
\texttt{hard\_check\_treatment},\ \texttt{hard\_check\_survival},\ \texttt{thinking}, \\
\texttt{cnlc},\ \texttt{bclc},\ \texttt{tnm},\ \texttt{answer\_treatment}, \texttt{answer\_survival}, 
\end{array}
\right\},
\] 
    where only \texttt{treatment} and \texttt{treatment\_not\_recommended} may appear multiple times. 
    The reward is defined as
    \begin{equation}
        R_{\text{fmt}} = \frac{1}{M} \sum_{m=1}^{M} s(\text{tag}_m),
    \end{equation}
    where $M = |\mathcal{T}_{\mathrm{req}}|$ and $s(\text{tag}_m)$ measures the correctness and uniqueness of each tag. 
    This encourages structured, machine-parsable outputs.

    \item \textbf{Treatment Ranking Reward} ($R_{\text{treat}}$).  
    Because multiple therapies can be clinically acceptable yet differ in expected benefit, we encourage higher ranks for guideline-consistent treatments using normalized discounted cumulative gain (nDCG). 
    First, define the discounted cumulative gain (DCG):
    \begin{equation}
    DCG = \sum_{i=1}^k \frac{\mathbb{I}(\hat{y}_i \in Y^*)}{\log_2(i+1)},
    \end{equation}
    where $\hat{y}_i$ is the $i$-th ranked treatment according to the predicted scores and $Y^*$ is the set of guideline-consistent treatments. 
    The normalized score is
    \begin{equation}
    R_{\text{treat}} = \frac{DCG}{IDCG},
    \end{equation}
    where $IDCG$ is the ideal DCG. 
    This ranking-based reward prioritizes the most clinically appropriate options.

    \item \textbf{Survival Reward} ($R_{\text{surv}}$).  
    Survival estimation is crucial for both staging and treatment planning. 
    This reward evaluates binary survival endpoints (1-, 3-, 5-year) and continuous month-level predictions. 
    For death-confirmed cases ($event=1$), we apply an exponential decay with respect to the absolute error:
    \[
    R_{\text{survival}} = \exp\big(-| \hat{t} - t | / \tau_{\text{death}} \big),
    \]
    where $\hat{t}$ is the predicted survival time, $t$ is the ground truth, and $\tau_{\text{death}}$ is a smoothing constant. 
    For censored cases ($\text{event}=0$), we penalize predictions earlier than the censoring bound $t_c$:
    \[
    R_{\text{survival}} = \exp\big(-\max(0, t_c - \hat{t}) / \tau_{\text{censor}} \big).
    \]
    To integrate both short-term and continuous measures, we define
    \begin{equation}
        R_{\text{surv}} = R_{\text{stage\_surv}} + R_{\text{month}},
    \end{equation}
    where the stage-wise accuracy is
    \begin{equation}
        R_{\text{stage\_surv}} = \frac{\sum_{k \in \{1,3,5\}} \mathbb{I}\{\hat{s}_k = s_k^{\text{GT}}, \ s_k^{\text{GT}} \in \{0,1\}\}}{\sum_k \mathbb{I}\{s_k^{\text{GT}} \in \{0,1\}\}},
    \end{equation}
    and the continuous term is
    \begin{equation}
        R_{\text{month}} = \exp\left(-\frac{|\hat{m} - m|}{\tau}\right), \quad \tau = 12,
    \end{equation}
    where $\hat{m}$ is the predicted survival months and $m$ the ground truth or censoring bound. 
    This design is consistent with standard survival analysis practice.

    \item \textbf{Length Reward} ($R_{\text{len}}$).  
    To discourage excessively verbose outputs, we impose a length regularization:
    \begin{equation}
        R_{\text{len}} = \min\left(1, \frac{L_{\max}}{L_{\text{pred}}}\right),
    \end{equation}
    where $L_{\text{pred}}$ is the sequence length and $L_{\max}$ a predefined threshold. 
    This enforces conciseness and clinical usability.
    
\end{enumerate}

By incorporating these components, the RL process went beyond token-level accuracy and promoted structured reasoning, interpretable predictions, and clinically valid recommendations. The overall composite reward is: 
\begin{equation}
    R(\hat{y}, y) =  R_{\text{proc}} + R_{\text{fmt}} +  R_{\text{treat}} +  R_{\text{surv}} +  R_{\text{len}}.
\end{equation}

\textbf{Group Relative Policy Optimization (GRPO).}
We adopted Group Relative Policy Optimization (GRPO)~\cite{shao_deepseekmath_2024}, a PPO-style~\cite{schulman_ppo_2017} method that aligned the policy with the composite reward without a value model. For each prompt $x$, we sampled $K$ candidates $\{\hat y_i\}_{i=1}^{K}$ from the behavior policy $\pi_{\text{old}}$ and computed scalar rewards $\{R_i\}$. To compare candidates under the same prompt, we formed a group-normalized advantage:
\begin{equation}
A_i=\frac{R_i-\mu_R}{\sigma_R+\epsilon_{\text{std}}},\qquad
\mu_R=\frac{1}{K}\sum_{j=1}^{K}R_j,\quad
\sigma_R^2=\frac{1}{K}\sum_{j=1}^{K}(R_j-\mu_R)^2,
\label{eq:grpo-adv}
\end{equation}
where $A_i$ was broadcasted to tokens for credit assignment, i.e., $A_{i,t}\!=\!A_i$ for all $t$ in sequence $\hat y_i$.

Following PPO, we used a token-level importance ratio
\begin{equation}
r_{i,t}(\theta)=
\frac{\pi_{\theta}\!\big(\hat y_{i,t}\mid x,\hat y_{i,<t}\big)}
     {\pi_{\text{old}}\!\big(\hat y_{i,t}\mid x,\hat y_{i,<t}\big)},
\label{eq:grpo-imp}
\end{equation}
and optimized a clipped surrogate averaged over candidates and tokens, with a KL penalty that constrained drift from a frozen reference policy $\pi_{\mathrm{ref}}$ (initialized from SFT):
\begin{equation}
\begin{aligned}
\mathcal{L}_{\text{GRPO}}(\theta)
= -\,\mathbb{E}\Bigg[
&\frac{1}{K}\sum_{i=1}^{K}\frac{1}{|\hat y_i|}\sum_{t=1}^{|\hat y_i|}
\min\!\Big(
r_{i,t}(\theta)\,A_{i,t},\;
\mathrm{clip}\big(r_{i,t}(\theta),\,1-\varepsilon,\,1+\varepsilon\big)\,A_{i,t}
\Big) \\
&\qquad -\ \beta\,D_{\mathrm{KL}}\!\big(\pi_{\theta}\,\|\,\pi_{\mathrm{ref}}\big)
\Bigg].
\end{aligned}
\label{eq:grpo-loss}
\end{equation}
This objective preserved PPO’s trust-region behavior through clipping, eliminated the need for a value model, and applied group-relative scaling to deliver a low-variance learning signal that reliably distinguished high- and low-quality generations for the same input. These properties were especially valuable in heterogeneous clinical tasks with censored outcomes and partial supervision.

\textbf{Decoupled optimization strategy.} 
To enhance training stability with heterogeneous objectives, we adopted a decoupled optimization strategy in which treatment-related and survival-related rewards were optimized in separate GRPO updates. Each reward produced an independent gradient, and we subsequently aggregated the updates during policy optimization. This approach mitigated gradient interference between tasks, preserved task-specific optimization dynamics, and still allowed the model to benefit from joint multi-objective reinforcement learning.

\subsection{Reasoning data synthesis}
\textbf{Alignment with clinical guidelines and expert consensus.}
We designed prompts in strict accordance with the \emph{Guidelines for the Diagnosis and Treatment of Primary Liver Cancer (CNLC, 2024 Edition)}~\cite{jia2024chinese} to support reinforcement learning with structured clinical reasoning. Each prompt instructed the model to act as a senior hepatobiliary clinical researcher, evaluate patient information, including performance status, Child--Pugh score, tumor burden, metastasis, and imaging findings, and provide step-by-step reasoning aligned with the guideline decision tree. The prompt design was provided in {Supplementary Listing~\ref{lst:hcc_prompt_thinking}}.  

To mimic real-world clinical decision-making, prompts were enumerated to encompass all potential treatment modalities, including surgical resection, ablation, liver transplantation, TACE, systemic therapy, radiotherapy, supportive care, and palliative care. The model was required to assign each option a continuous suitability score between 0 and 1, accompanied by a justification consistent with the guideline decision tree. We further aligned the chain-of-thought reasoning with expert consensus by requiring outputs to explicitly reference \emph{indications, contraindications, levels of evidence, and recommendation grades}. Beyond listing feasible therapies, the model was asked to compare guideline-recommended options and articulate the rationale for ranking them. For instance, in early-stage single tumors ($\leq$3\,cm), resection was prioritized over ablation based on reported survival benefits (evidence level~1, recommendation~A), while transplantation was downgraded due to donor scarcity and fairness considerations (evidence level~3, recommendation~B).

\textbf{Structured and verified tagging.}
We enforced structured tagging to facilitate automatic evaluation during reinforcement learning, and a specific prompt was provided in {Supplementary Listing~\ref{lst:tag_en_full}}. Clinical variables such as PS, Child--Pugh score, tumor size, number of tumors, metastasis status, and vascular invasion were annotated with XML-style tags (e.g., \texttt{<ps>}, \texttt{<child\_pugh>}, \texttt{<tumor\_size>}). This ensured both reliable parsing for reward assignment and transparency for clinicians, who could verify that model outputs referenced appropriate clinical evidence. 

Survival-related fields, including cause of death and survival months, were included in the input but explicitly masked during generation. The model was required to predict survival status and provide justification without directly referencing ground-truth outcomes, consistent with standard survival analysis settings involving censoring~\cite{kaplan_meier_1958,cox1972}. This allowed the same dataset to support both \textit{treatment recommendation rewards} and \textit{survival prediction rewards}, forming the basis for supervised fine-tuning and reinforcement learning.

\textbf{Scoring and reward computation.}
At each stage, the model evaluated candidate treatments using \emph{continuous preference scores}. For a patient case $x$, the output was represented as:
\begin{equation}
    \hat{\mathbf{s}} = (\hat{s}_1, \hat{s}_2, \ldots, \hat{s}_M), \quad \hat{s}_j \in [0,1],
\end{equation}
where $M$ denotes the number of candidate treatments, and $\hat{s}_j$ reflects the predicted suitability of treatment $j$. Ground-truth scores $\mathbf{s}^\ast$ were derived from guideline-consistent expert consensus. Each output contained:  
\begin{enumerate}
    \item A \texttt{<thinking>} section with explicit reasoning steps referencing patient-specific features.  
    \item Structured tags for intermediate diagnostic variables (e.g., \texttt{<ps>}, \texttt{<child\_pugh>}, \texttt{<metastasis>}, \texttt{<cancer\_thrombus>}, \texttt{<num\_tumor>}, \texttt{<tumor\_size>}).  
    \item Treatment annotations using \texttt{<treatment>} and \texttt{<treatment\_not\_recommended>} tags.  
    \item A \texttt{<hard\_check>} section containing a JSON dictionary of treatment scores $\hat{s}_j$, enabling quantitative evaluation.  
\end{enumerate}

This structured design allowed direct computation of reward signals. For example, the treatment-ranking reward was obtained by comparing $\hat{\mathbf{s}}$ with $\mathbf{s}^\ast$ using normalized discounted cumulative gain (nDCG). The explicit reasoning chain further supported \emph{process rewards} by verifying intermediate variables. By combining continuous scoring, structured outputs, and interpretable reasoning, the framework closely coupled guideline-consistent decision making with reinforcement-learning objectives and ensured both reliability and trainability under the GRPO algorithm.

\subsection{Comparison with competitive baselines}
To rigorously assess performance in treatment recommendation and survival prognosis, we designed two complementary evaluations. First, we compared the proposed LLM-based system with traditional machine-learning (ML) methods and with open-source and commercial LLMs. Second, we benchmarked prognostic discrimination against widely used staging systems (AJCC (TNM), BCLC, and CNLC) that reflect different clinical guidelines. We measured treatment recommendation using Top-$k$ accuracy, and we evaluated survival prognosis using time-dependent Harrell’s concordance index (C-index) and ROC curves across multiple horizons, together with Kaplan–Meier analyses for risk stratification.

For the ML baselines, we implemented representative classical, probabilistic, and neural approaches, including support vector machines (SVMs)~\cite{cortes1995svm}, XGBoost~\cite{chen2016xgboost}, Bayesian models~\cite{van2021bayesian}, and multilayer perceptrons (MLPs)~\cite{rumelhart1986backprop}. These models were trained on structured variables extracted from SEER (e.g., laboratory markers, tumor size, vascular invasion, and extrahepatic metastasis) to jointly predict treatment categories and survival time. Because traditional ML methods depend on manually abstracted features and fixed variable schemas, these baselines cannot generalize to the external multi-center cohort, where formats and variable availability differ from SEER. Consequently, comparisons with ML methods were restricted to the internal SEER test set. In contrast, our LLM directly processed unstructured EMR narratives, preserved patient-level detail, and captured nonlinear interactions among tumor burden, liver function, vascular invasion, and treatment patterns. The LLM also generated explicit chain-of-thought reasoning that provided transparent support for clinical interpretation.

\subsection{Metrics}
We measured treatment recommendation using Top-$k$ accuracy, and we assessed prognosis with Harrell’s concordance index (C-index) and time-dependent area under the ROC curve (AUROC) at 1, 3, and 5 years. Risk stratification was quantified with Kaplan–Meier analyses using model-predicted stagings and prespecified low/high-risk cutoffs.

\textbf{Top-$k$ accuracy.} We introduced a \emph{rank-aware} Top-$k$ similarity that quantifies both list overlap and ordering concordance. For each patient \(x\), let \(\hat T_k(x)=(\hat t_1,\ldots,\hat t_k)\) denote the Top-\(k\) \emph{predicted} list (sorted by decreasing model score) and \(T_k(x)=(t_1,\ldots,t_k)\) the Top-\(k\) \emph{ground-truth} list (sorted by decreasing ground-truth score). Higher scores correspond to earlier ranks. Ground-truth scores were thresholded at \(\tau=0.5\). Items below the threshold were excluded, and if fewer than \(k\) items remained, missing positions contributed zero (i.e., no overlap term). We define the rank-aware Top-$k$ similarity as
\begin{equation}
\mathrm{Sim}_k(x)
= \frac{1}{k}\sum_{i=1}^{k}\sum_{j=1}^{k}
\mathbb{I}\!\left[\hat t_i = t_j\right]\cdot \frac{1}{1+\lvert i-j\rvert}.
\label{eq:rankaware-topk}
\end{equation}
Equivalently, writing \(\widehat{\mathrm{pos}}(u)=\min\{i:\hat t_i=u\}\), \(\mathrm{pos}(u)=\min\{j:t_j=u\}\), and \(\mathcal T=\{\,u \mid u\in \hat T_k(x)\cap T_k(x)\,\}\),
\begin{equation}
\mathrm{Sim}_k(x)
= \frac{1}{k}\sum_{u\in \mathcal T}
\frac{1}{1+\big\lvert \widehat{\mathrm{pos}}(u)-\mathrm{pos}(u)\big\rvert}.
\label{eq:rankaware-topk-intersection}
\end{equation}
By construction, \(\mathrm{Sim}_k(x)\in[0,1]\), it equals 1 only when the Top-\(k\) items match exactly in membership and order, and it decreases as overlap shrinks or ranks diverge.

\textbf{Prognosis stratification:} Harrell’s C-index measures the proportion of correctly ordered patient pairs with respect to observed survival and censoring, which reflects global ranking performance over follow-up. Time-dependent AUROC at 1, 3, and 5 years quantifies horizon-specific discrimination by treating the model’s continuous risk as a score and evaluating case–control separability among those at risk at each time point. We reported the overall C-index and horizon-specific C-index where indicated, together with AUROC values at the same horizons.

\textbf{Kaplan–Meier:} We generated survival curves by \emph{model-predicted staging} (A–D) using quantile-based thresholds learned on the training cohort and transferred without change to the test cohorts, and by \emph{risk group} using a prespecified low/high cutoff. Curves summarize event trajectories over time and support comparisons of median survival and group separation.

\textbf{Hypothetical OS analyses under alternative treatment recommendations:}
To compare treatment algorithms at the cohort level, we constructed hypothetical overall-survival curves by assigning each patient to the therapy recommended by a given algorithm (either our model or a clinical staging scheme). For patients who did not receive the recommended therapy, we imputed outcomes by resampling from clinically similar cases within the test set who did receive that therapy. Kaplan–Meier curves were then estimated, and median OS under each algorithm was reported to reflect potential differences in population-level outcomes.

\subsection{Study design and participants for clinical evaluation}
We designed a comprehensive clinical study that incorporates three components: (i) a human expert evaluation framework to assess the quality of LLM-generated content, (ii) a direct comparison of diagnostic accuracy between the LLM system and physicians at different levels of seniority, and (iii) an assessment of the model’s effectiveness in assisting junior and intermediate physicians in treatment decision-making. A panel of 13 hepatobiliary physicians with varying levels of clinical experience is recruited, comprising three junior (resident) physicians with 1–5 years of practice experience, four intermediate (attending) physicians with 5–10 years of experience, and six senior physicians with more than 10 years of experience. This study was approved by the Beijing Tsinghua Changgung Hospital Medical Science Research Ethics Committee (IRB 25532-4-01).

\textbf{The evaluation of the LLM-generated content.}
To rigorously evaluate both the CoT and the plausibility of the final treatment category, we convened three senior hepatobiliary specialists to develop a consensus-based questionnaire aligned with clinical practice (Supplementary Section~\ref{sec:treat_eval}). The same three senior specialists then assembled a retrospective test set of 60 HCC cases, manually curated rather than randomly sampled, to span (i) BCLC, CNLC, and AJCC/TNM stages from early to advanced disease, (ii) a range of clinical difficulty including borderline and contraindication-laden profiles, and (iii) the major candidate-therapy categories (resection, ablation, transplant referral, TACE, systemic therapy, radiotherapy, supportive or palliative care). Baseline characteristics of the resulting cohort are summarized in Extended Data Fig.~\ref{fig:60_cases_stats}.

We then assembled a retrospective test set of 60 HCC cases spanning different levels of difficulty and candidate therapies. The outputs of three models (our system, DeepSeek-v3, and GPT-4o) are randomly mixed and presented to physicians for blinded scoring, with all case details and model identities de-identified. For each case, three specialists (each with at least 15 years of clinical experience) independently provide a reference for the first-line ranked treatment, and the majority decision (two out of three) serves as the reference standard. Then, three senior physicians reviewed and scored the anonymized CoT and treatment recommendations using the designed questionnaire, which evaluates \emph{completeness} (see Supplementary Table~\ref{tab:completeness}), \emph{correctness}, and \emph{safety} (see Supplementary Table~\ref{tab:safety}). The composite quality score (QNS) is calculated as the sum of these three domains. In addition, to evaluate the soundness of the evidence-based justification, three specialists also assessed the supporting evidence for each recommended treatment across three dimensions: \emph{completeness}, \emph{correctness} and \emph{consistency}. 

\textbf{Comparison of treatment accuracy between our model and physicians.}
Using the same blinded case set (60 heterogeneous HCC cases), we compared our model with junior and intermediate hepatobiliary physicians and two LLMs on Top-$k$ treatment accuracy. 
Each reader formulated treatment recommendations from de-identified EMR-derived clinical vignettes that included demographics, chief complaint, history of present illness, prior therapies, physical examination findings, laboratory data, and imaging reports. Model identities were concealed and case order was randomized.
A reference standard was established by an expert consensus panel of three senior hepatobiliary oncologists, who independently proposed guideline-consistent treatment sets with ranked suitability scores and then resolved any disagreement through adjudication. The resulting consensus list served as the ground truth for Top-$k$ accuracy measuring. We also recorded whether any option violated absolute or practical contraindications, which contributed to the safety-pass metric reported in the Results.

\textbf{Assisted treatment decision-making with our LLM-based system.}
We conducted a study to examine the potential role of our system in assisting physicians during treatment planning. After the initial session without assistance, each junior and intermediate physician completed a second session in which the model’s output was displayed alongside the same clinical vignette, including the ranked treatment options, concise reasoning rationales, and the tagged intermediate determinants such as \texttt{<ps>}, \texttt{<child\_pugh>}, and \texttt{<metastasis>}. The assisted session was scheduled at least two weeks after the baseline session to mitigate recall. During this session, physicians were free to accept, reorder, or override the model’s suggestions and then submit a final ranked list. We compared Top-$k$ accuracy and decision time before and after assistance and benchmarked the assisted performance against the standalone model and the senior-expert consensus, which allowed us to assess whether integrating the LLM into the workflow improved decision quality for junior and intermediate physicians.

\section{Statistical analysis}
We report two-sided 95\% confidence intervals (CIs) estimated by patient-level bootstrap unless stated otherwise. Pairwise C-index comparisons between our model and baselines use two-sided tests with $P$ values reported in figures and tables. Time-dependent ROC comparisons use DeLong’s test or IPCW-based variants when appropriate. For Kaplan–Meier analyses, groups are compared with the two-sided log-rank test. Median survival and 95\%~CIs are reported using standard KM estimators (Greenwood-based errors; Brookmeyer–Crowley CIs for medians when applicable). Hazard ratios (HRs) and 95\%~CIs are approximated from proportional-hazards models. In clinical studies, a two-sided $P$ value of less than 0.05 is considered statistically significant. 

\section{Data availability}\label{sec5}
The Surveillance, Epidemiology, and End Results (SEER) program data used in this study are publicly available from the National Cancer Institute. Researchers can request access and download SEER files via the SEER website and SEER*Stat portal, in accordance with the SEER data-use agreement and citation guidance (\url{https://seer.cancer.gov/}). 
The multi-center electronic medical records (EMRs) assembled for this work comprise de-identified but sensitive real-world clinical data from 12 tertiary hospitals across China. In compliance with institutional review board approvals and local regulations on patient privacy, these data cannot be deposited in a public repository. De-identified EMRs can be made available for non-commercial, academic research upon reasonable request to the responsible authors (\textit{cui-peng@mail.tsinghua.edu.cn} and \textit{dongjiahong@mail.tsinghua.edu.cn}), subject to approval by the participating institutions. Requests should include a brief research proposal, proof of ethics approval (or exemption) from the requester’s institution, and a signed data-use agreement specifying secure handling, no re-identification, and no redistribution. Requests are typically reviewed within 20 working days; if approved, time-limited read access (e.g., 12 months) will be granted under the terms of the agreement.

Aggregate data underlying the figures and tables (for example, C-index, AUROC values, and Kaplan–Meier estimates) are provided as Source Data with this paper.

\section{Code availability}\label{sec6}
All deep learning experiments were developed in Python (3.12) with PyTorch (2.7.0). We used the following standard libraries: NumPy~2.1.2, pandas~2.3.1, SciPy (1.16.1), scikit-learn (1.2.1), Transformers (4.53.0), vLLM (0.9.2), and Matplotlib (3.10.5). Transformers provided tokenizer/model interfaces, scikit-learn and SciPy were used for metric computation and statistical testing, pandas and NumPy handled data preprocessing, and Matplotlib was used for figure generation; vLLM was used for efficient inference and serving. Supervised fine-tuning (SFT) and reinforcement learning were implemented on PyTorch (2.7.0) and Verl (0.5.0).
Specifically, we implemented {Knowledge-Aligned and Experience-Accrual reasoning} to optimize the composite reward for treatment recommendation and survival prediction, building on the SFT-initialized backbone. 
The codes are available for scientific research and non-commercial use on GitHub at \url{https://github.com/Aries-iai/HCC-STAR}.

\clearpage

\bibliographystyle{unsrt}
\bibliography{sn-bibliography}

\clearpage
\appendix
\section{Extended Data figures and tables}
\setcounter{figure}{0}  
\setcounter{table}{0}   

\renewcommand{\figurename}{Extended Data Fig.}
\renewcommand{\tablename}{Extended Data Table}

\begin{figure}[ht]
    \centering
    \includegraphics[width=0.95\linewidth]{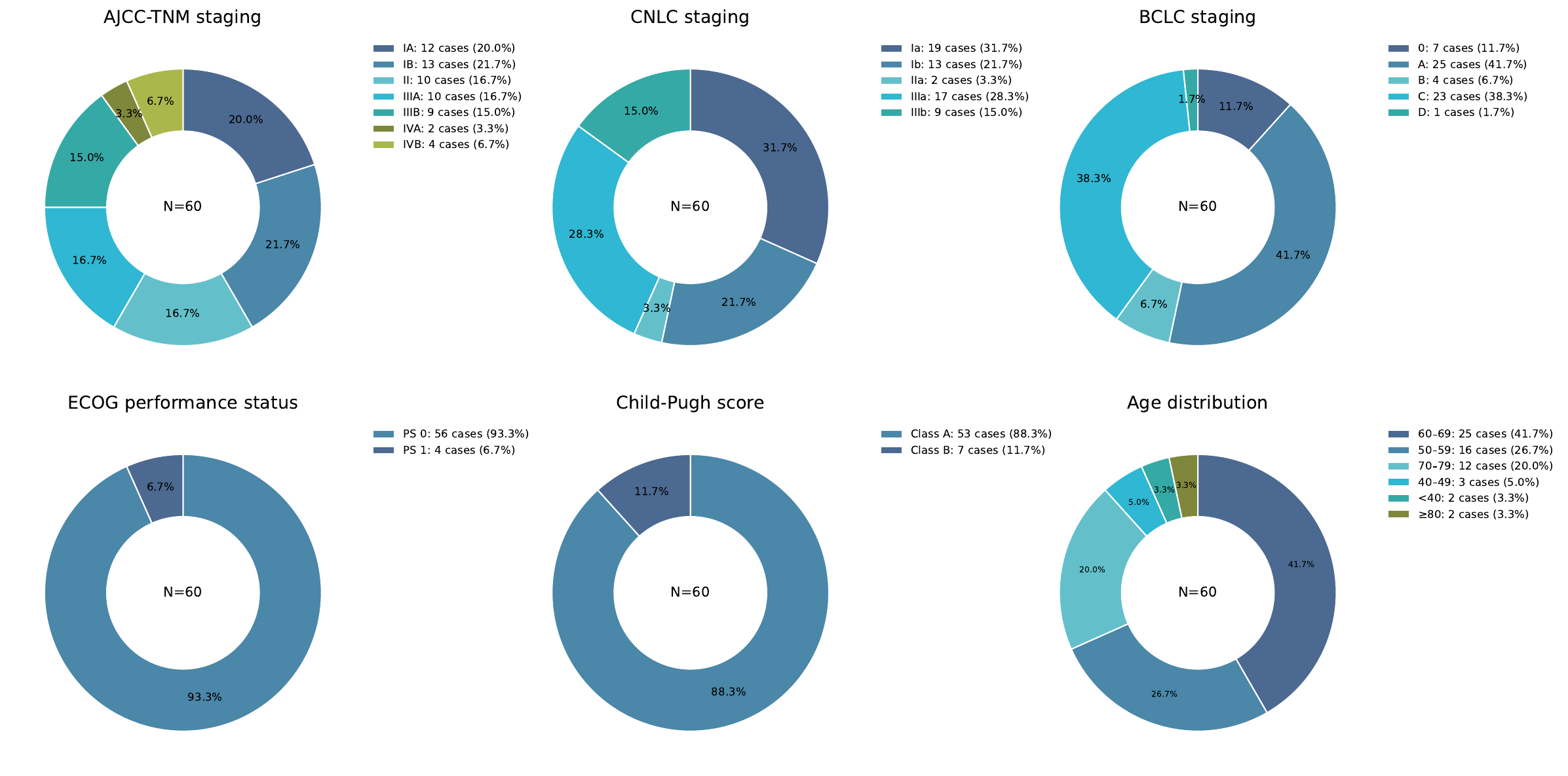}
    \caption{Baseline characteristics of the curated 60-case HCC cohort. Pie charts summarize the distributions of (top row) AJCC--TNM staging, CNLC staging, and BCLC staging, and (bottom row) ECOG performance status, Child--Pugh score, and age groups. Slice labels report the number of cases and the corresponding proportion of the cohort.}
    
    \label{fig:60_cases_stats}
\end{figure}

\begin{figure}[ht]
    \centering
    \includegraphics[width=0.99\linewidth]{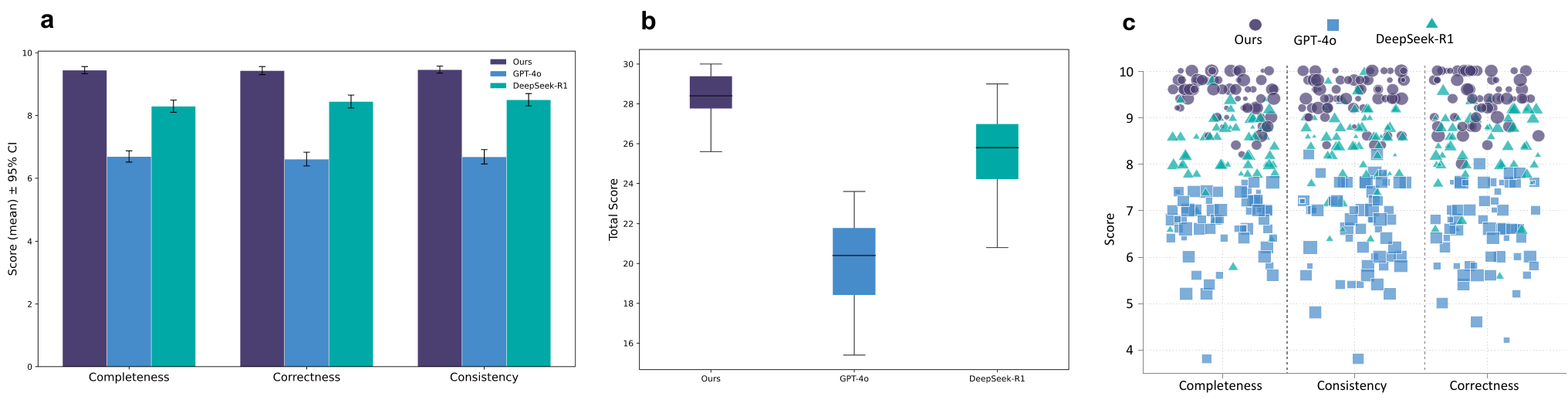}
    \caption{Blinded expert scoring of evidence-based justifications for our model and other LLMs. \textbf{a,} Three senior hepatobiliary specialists rated completeness, correctness, and consistency of evidence. \textbf{b,} Sum score of the three domains. \textbf{c,} Score distribution of three LLMs across 60 cases.}
    \label{fig:evidence_scores}
\end{figure}

\begin{figure}[H]
    \centering
    \includegraphics[width=0.98\linewidth]{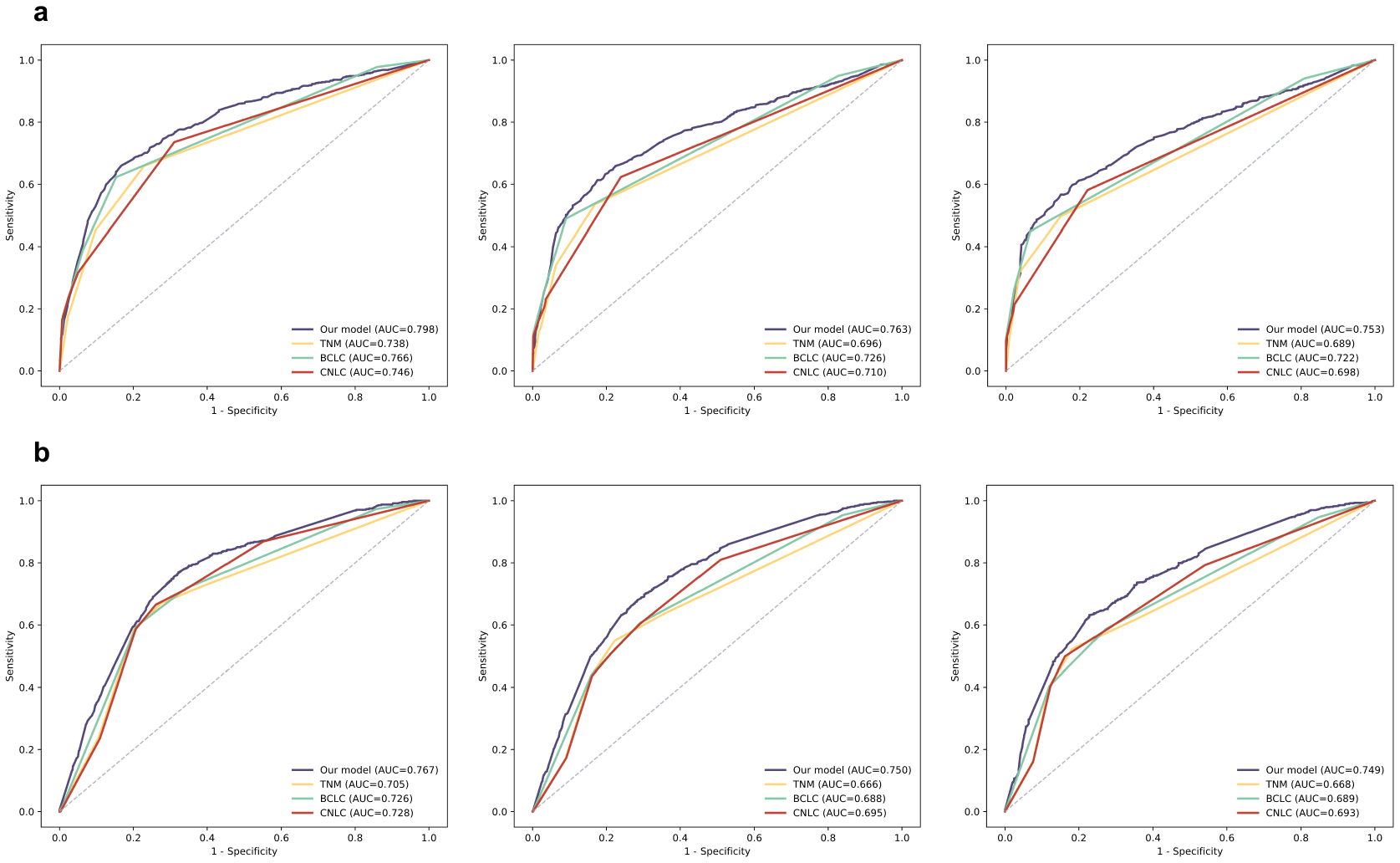}
    \caption{Time-dependent ROC: our method vs. clinical staging systems on the internal and external cohorts. \textbf{a}, Internal test cohort with ROC curves at 1, 3, and 5 years. \textbf{b}, External multi-center cohort with ROC curves at 1, 3, and 5 years. Curves quantify overall-survival risk discrimination at each horizon. Area under the ROC (AUROC) estimates and 95\% confidence intervals are summarized in Extended Data Table~\ref{tab:cindex_auc_internal} and Table~\ref{tab:cindex_auc_external}; two-sided DeLong $P$ values are reported alongside.}
    \label{fig:rocs_all}
\end{figure}

\begin{figure}[ht]
    \centering
    \includegraphics[width=\linewidth]{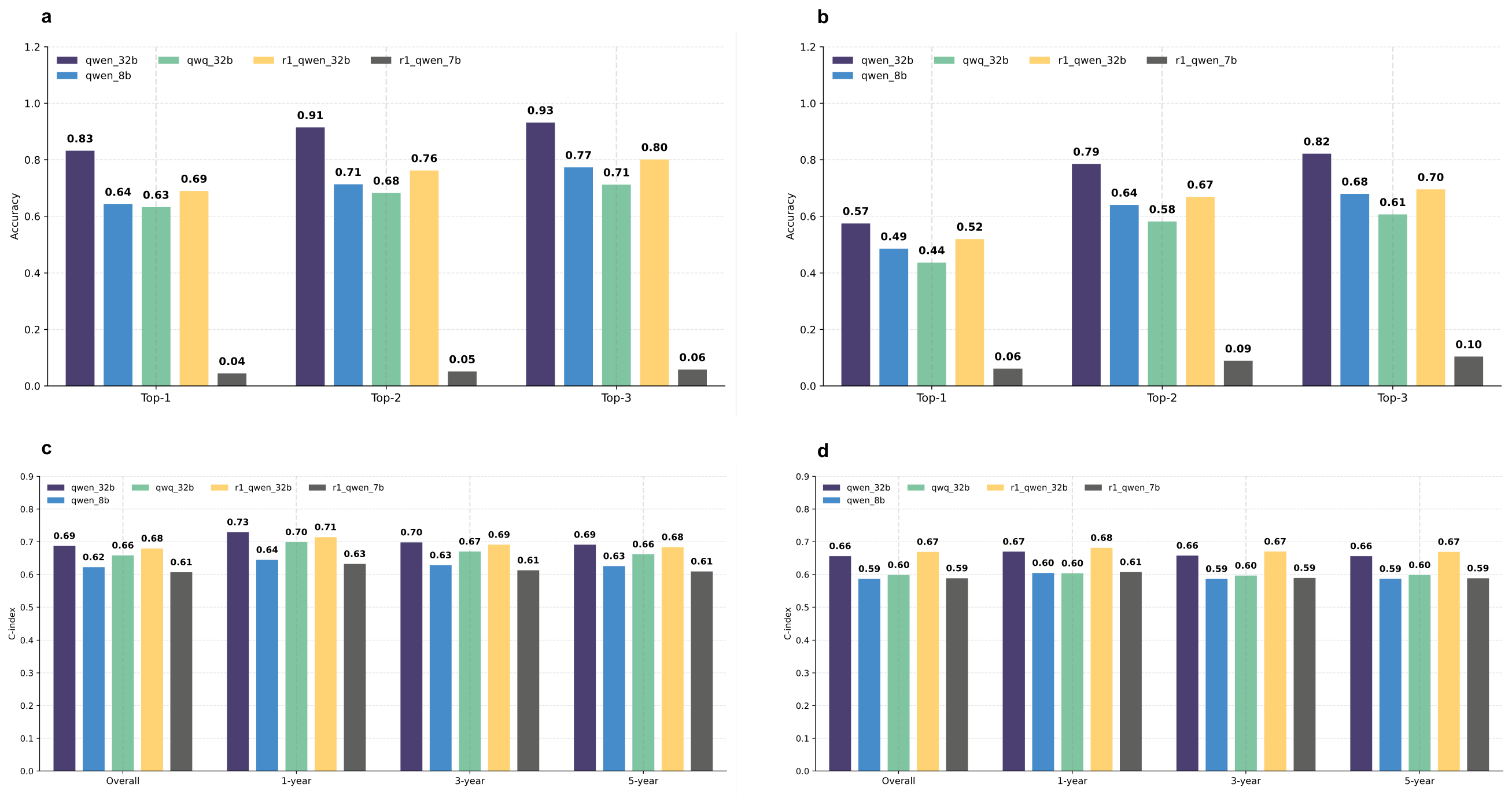}
    \caption{
\textbf{Supervised fine-tuning (SFT) performance for treatment accuracy and Concordance index (C-index) across LLM backbones.}
\textbf{a, b}, Top-$k$ treatment recommendation accuracy on the internal (a) and external (b) test cohorts. 
Each model outputs a ranked list of candidate therapies, and accuracy-$k$ counts a correct prediction when the reference treatment appears within the top-$k$ positions ($k = 1,2,3$).
\textbf{c, d}, C-index for overall, 1-year, 3-year, and 5-year survival prediction on the internal (c) and external (d) cohorts. 
Across metrics and cohorts, \textit{Qwen3-32B} consistently achieves the highest SFT performance among all evaluated backbones, motivating its use as the foundation for subsequent reinforcement-learning (RL) optimization in our full training pipeline.
}
    \label{fig:sfts_compare}
\end{figure}

\begin{figure}[ht]
    \centering
    \includegraphics[width=0.80\linewidth]{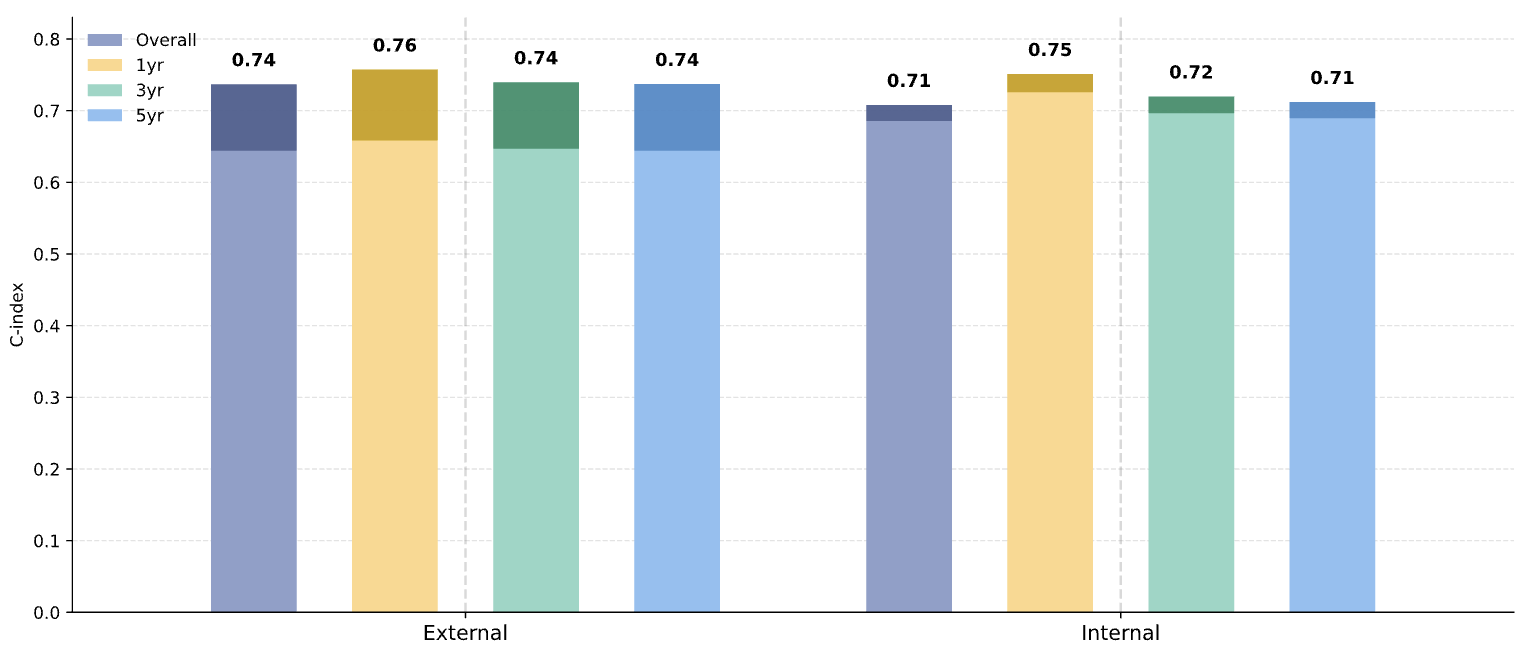}
    \caption{C-index with and without reinforcement learning (RL). Harrell’s C-index for supervised fine-tuning (SFT) and SFT+RL on the internal SEER test set and the external multi-center cohort, evaluated overall and at 1/3/5 years. Dark segments denote the incremental gain of C-index from RL. RL yielded especially pronounced gains on the external test set, demonstrating strong generalization.}
    \label{fig:cindex_sft_rl}
\end{figure}
\clearpage
\begin{figure}[H]
    \centering
    \includegraphics[width=0.99\linewidth]{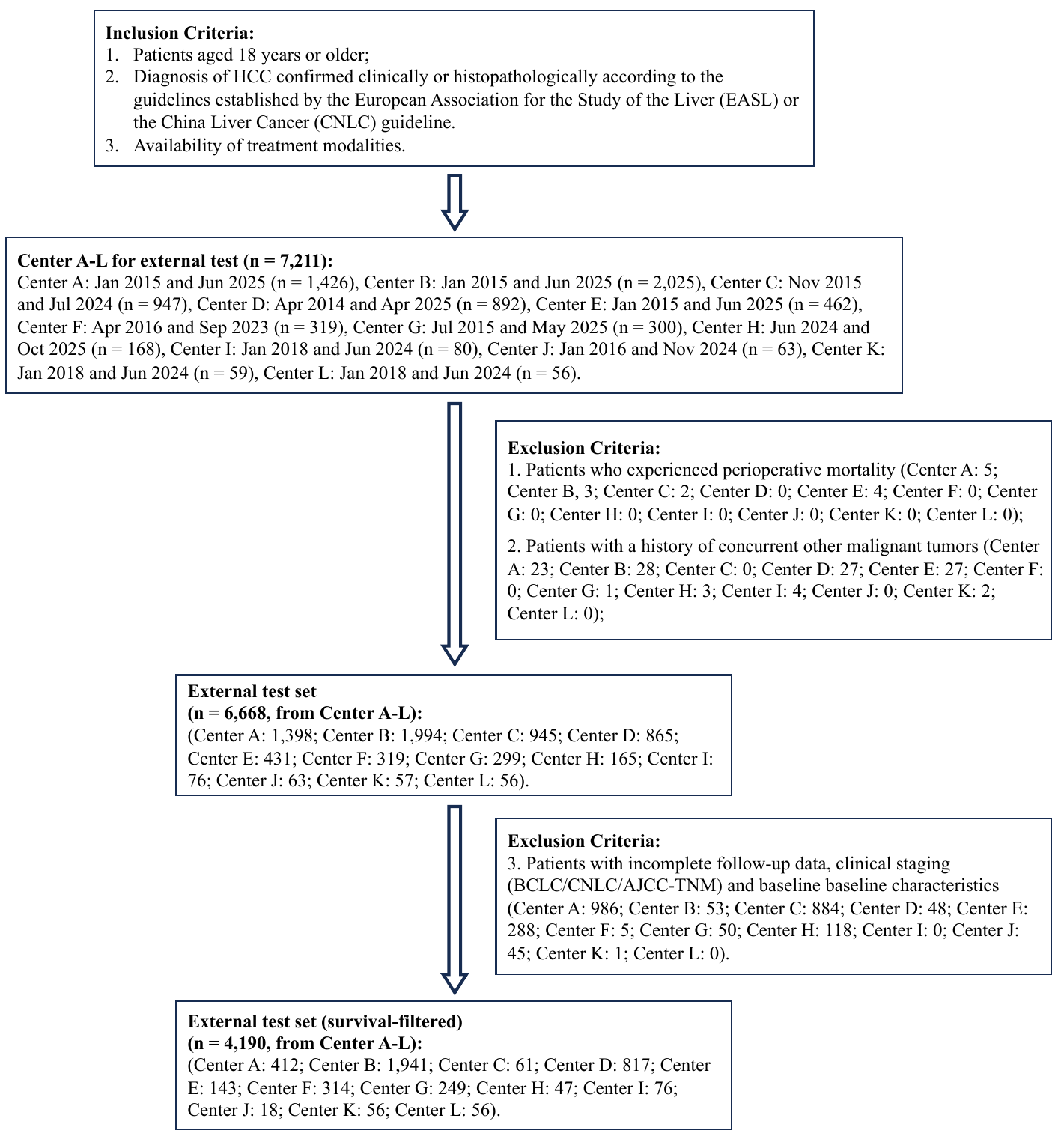}
    \caption{Flowchart of patient screening, exclusions, and final inclusion across twelve hospitals in China for the external validation cohort.}
    \label{fig:multi_center_flow}
\end{figure}

\begin{figure}[ht]
    \centering
    \includegraphics[width=0.99\linewidth]{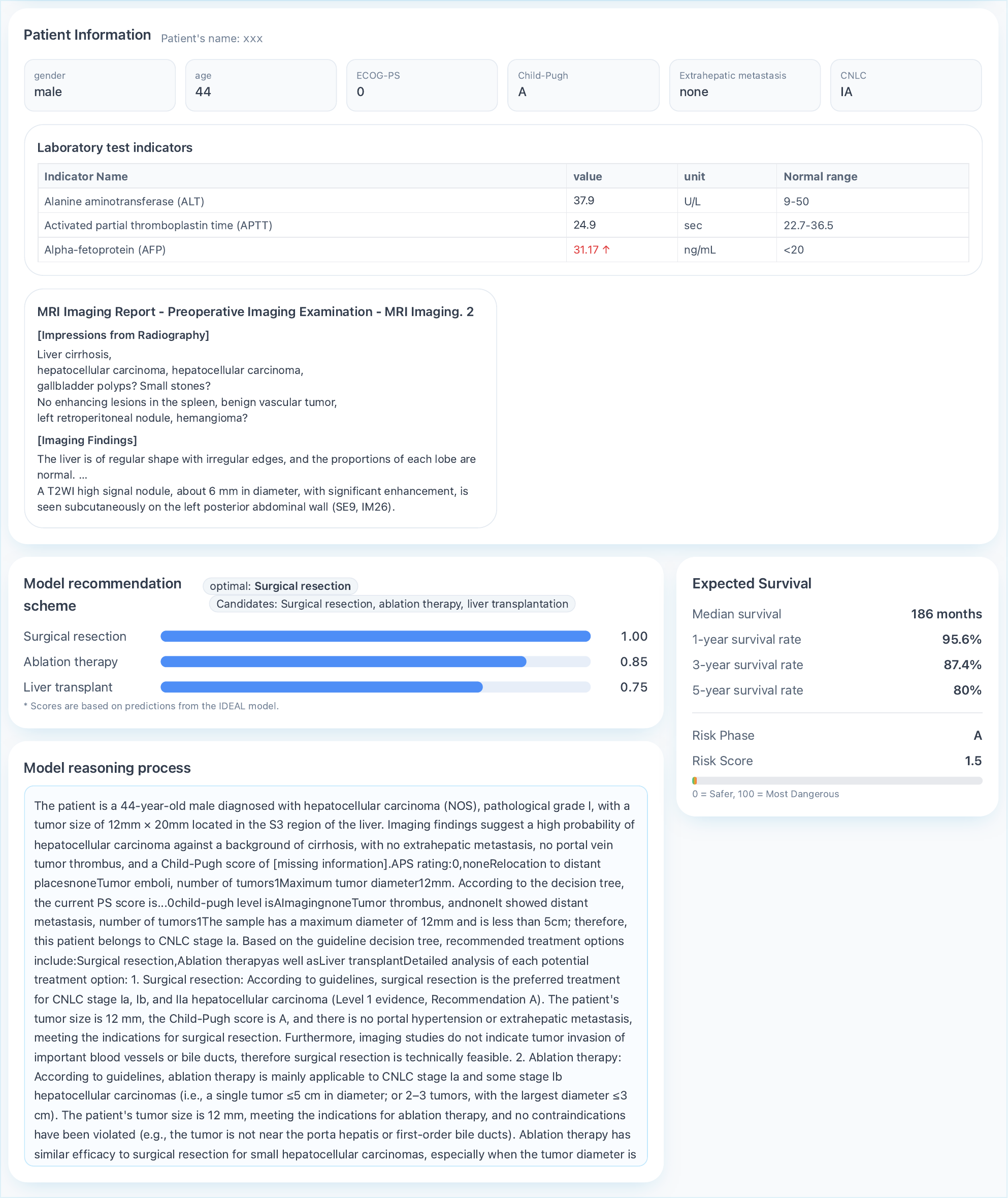}
    \caption{Our AI-assisted decision-making system of HCC for doctors.}
    \label{fig:sys_demo}
\end{figure}

\begin{figure}[ht]
    \centering
    \includegraphics[width=0.99\linewidth]{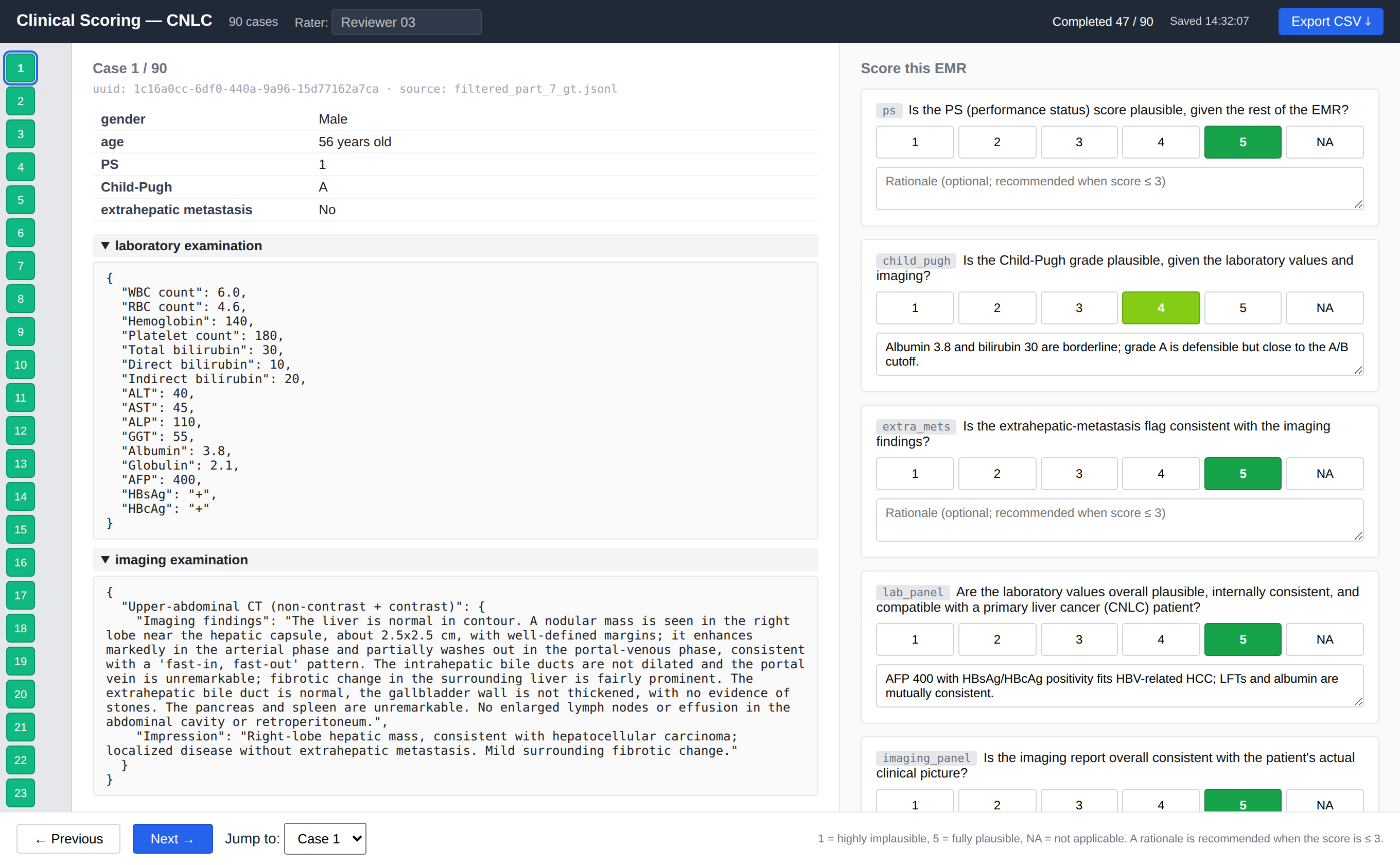}
    \caption{Web-based clinician scoring interface for record-level plausibility assessment of the CNLC hepatocellular carcinoma cohort.}
    \label{fig:cnlc_demo}
\end{figure}

\begin{table}[htbp]
\centering
\small
\setlength{\tabcolsep}{1.8pt}
\renewcommand{\arraystretch}{1.15}
\caption{Blinded clinical fidelity assessment of synthetic EMR narratives.}
\label{tab:synthetic_emr_fidelity}
\begin{threeparttable}
\begin{tabular}{p{3.3cm}ccccp{1.9cm}p{1.9cm}p{2.0cm}}
\toprule
\textbf{Assessment dimension} & \textbf{EMRs,n} & \textbf{Available ratings, n} & \textbf{Mean $\pm$ s.d.} & \textbf{Median [IQR]} & \textbf{Ratings$\geq$4, n(\%)} & \textbf{Ratings$\leq$2, n(\%)} & \textbf{EMRs mean $\geq$4, n(\%)} \\
\midrule
ECOG PS plausibility & 140 & 420 & 4.68 $\pm$ 0.68 & 5 [5--5] & 387 (92.1) & 7 (1.7) & 126 (90.0) \\
Child--Pugh plausibility & 140 & 420 & 4.72 $\pm$ 0.66 & 5 [5--5] & 393 (93.6) & 7 (1.7) & 129 (92.1) \\
Extrahepatic metastasis--imaging consistency & 140 & 420 & 4.92 $\pm$ 0.43 & 5 [5--5] & 413 (98.3) & 5 (1.2) & 137 (97.9) \\
Laboratory panel coherence & 140 & 420 & 4.93 $\pm$ 0.27 & 5 [5--5] & 418 (99.5) & 0 (0.0) & 139 (99.3) \\
Imaging-report realism & 140 & 418 & 4.86 $\pm$ 0.46 & 5 [5--5] & 413 (98.8) & 4 (1.0) & 137 (97.9) \\
Global narrative consistency & 140 & 420 & 4.73 $\pm$ 0.64 & 5 [5--5] & 394 (93.8) & 8 (1.9) & 129 (92.1) \\
\midrule
Overall & 140 & 2,518 & 4.81 $\pm$ 0.55 & 5 [5--5] & 2,418 (96.0) & 31 (1.2) & 137 (97.9) \\
\bottomrule
\end{tabular}
\begin{tablenotes}[flushleft]
\footnotesize
\item Scores used a five-point Likert scale, where 1 indicated highly unreasonable and 5 indicated fully reasonable; NA was allowed for non-applicable fields and excluded from denominators. Three senior hepatobiliary clinicians independently evaluated all 140 synthetic EMRs across six dimensions. Across 839 complete case--dimension triplets, three-rater exact agreement was 635/839 (75.7\%) and three-rater within-one-point agreement was 799/839 (95.2\%). Pairwise exact agreement was 83.0\%, and pairwise within-one-point agreement was 97.1\%. The overall ICC(2,k) was 0.73 and Krippendorff's $\alpha$ was 0.47.
\end{tablenotes}
\end{threeparttable}
\end{table}

\begin{table}[t]
    \centering
    \renewcommand\arraystretch{1.1}
    \setlength{\tabcolsep}{2.2mm}
    
    \caption{Comparisons between our model and major guideline staging systems and LLMs on the external multi-center cohort. Classical ML baselines are reported on the internal cohort only because the external dataset uses a different schema and variable format. (Data in parentheses are the 95\% confidence intervals. P values were calculated by comparing with our model using the two-sided DeLong test, and P $<$ 0.05 indicated a significant difference.)}
\begin{tabular}{lcccc}
    \toprule
        Models & C-index & P value & AUROC & P value \\ \midrule
        Our model & \textbf{0.7371 (0.7234–0.7507)} & ~ & \textbf{0.7573 (0.7193–0.7953)} & ~ \\ 
        AJCC staging system & 0.6594 (0.6442–0.6754) & $<0.001$ & 0.6514 (0.6152–0.6876) & $<0.001$ \\ 
        BCLC staging system & 0.6789 (0.6639–0.6940) & $<0.001$ & 0.6719 (0.6347–0.7090) & $<0.001$ \\ 
        CNLC staging system & 0.6846 (0.6695–0.6996) & $<0.001$ & 0.6808 (0.6426–0.7189) & $<0.001$ \\ \midrule
        GPT-4o & 0.7002 (0.6856–0.7146) & $<0.001$ & 0.6555 (0.6158–0.6952) & $<0.001$ \\
        GPT-5 & 0.7216 (0.7071–0.7372) & $0.136$ & 0.6534 (0.6129–0.6940) & $<0.001$ \\ 
        Gemini-2.5pro & 0.7274 (0.7124–0.7425) & $0.304$ & 0.6692 (0.6290–0.7095) & $<0.001$ \\ 
        Deepseek-R1 & 0.7201 (0.7050–0.7347) & $0.084$ & 0.6501 (0.6097–0.6905) & $<0.001$ \\ 
        Claude & 0.7186 (0.7048–0.7330) & $0.074$ & 0.6498 (0.6095–0.6901) & $<0.001$ \\ \bottomrule
\end{tabular}
    \label{tab:cindex_auc_external}
\end{table}

\begin{table}[htbp]
    \centering
    \renewcommand\arraystretch{1.1}
    \setlength{\tabcolsep}{2.2mm}
    
    \caption{Comparisons between our model and three guideline staging systems and machine learning methods on the internal testing cohort (Data in parentheses are the 95\% confidence intervals. P values were calculated by comparing with our model using the two-sided DeLong test, and P $<$ 0.05 indicated a significant difference.)}
    \begin{tabular}{lcccc}
    \toprule
        Models & C-index & P value & AUROC & P value \\ \midrule
        Our model & \textbf{0.7079 (0.6935–0.7228)} & ~ & \textbf{0.7325 (0.6634–0.8016)} & ~ \\ 
        AJCC staging system & 0.6581 (0.6438–0.6717) & $<0.001$ & 0.6611 (0.5987–0.7235) & $0.001$ \\ 
        BCLC staging system & 0.6808 (0.6680–0.6942) & $<0.001$ & 0.6711 (0.6062–0.7360) & $0.004$ \\ 
        CNLC staging system & 0.6659 (0.6524–0.6794) & $<0.001$ & 0.6586 (0.5905–0.7268) & $0.004$ \\ \midrule
        XGBoost & 0.7003 (0.6853–0.7147) & $0.009$ & 0.7189 (0.6500–0.7878) & $0.148$ \\
        MLP & 0.6830 (0.6688–0.6961) & $<0.001$ & 0.6619 (0.5969–0.7270) & $0.010$ \\ 
        SVM & 0.6670 (0.6525–0.6806) & $<0.001$ & 0.6020 (0.5337–0.6703) & $<0.001$ \\ 
        Bayesian models & 0.6398 (0.6268–0.6523) & $<0.001$ & 0.4907 (0.4250–0.5564) & $<0.001$ \\ \bottomrule
    \end{tabular}
    \label{tab:cindex_auc_internal}
\end{table}

\begin{table}[t]
\renewcommand{\arraystretch}{1.25} 
\centering
\caption{SEER variables used in this study, grouped by category.}
\label{tab:seer_fields_grouped}
\begin{tabular}{p{6cm} p{7.5cm}}
\toprule
\textbf{Field} & \textbf{Description} \\
\midrule
\multicolumn{2}{l}{\textbf{Patient demographics}} \\
Sex & Patient's biological sex \\
Age & Age at initial diagnosis \\
Marital.status.at.diagnosis & Marital status at diagnosis \\
\hline
\multicolumn{2}{l}{\textbf{Tumor characteristics}} \\
Grade & Tumor differentiation / histological grade \\
ICD-O-3.Hist.behav & Histology and behavior code (ICD-O-3 standard) \\
Stage & SEER summary stage at diagnosis \\
T / N / M & Tumor, nodal, and metastasis categories\\
Tumor.size & Maximum tumor size (mm) \\
AFP & Alpha-fetoprotein (AFP) biomarker status \\
Fibrosis & Fibrosis score (e.g., Ishak or bridging fibrosis) \\
Mets.bone / brain / liver / lung / other & Indicators for specific metastasis sites \\
First.malignant.primary.indicator & Whether this is the first malignant primary tumor \\
Total.number.of.in.situ.malignant.tumors & Number of in situ or malignant tumors recorded \\
\hline
\multicolumn{2}{l}{\textbf{Treatment information}} \\
Surgery & Indicator for surgical treatment received \\
Surg.lymn1 & Number of regional lymph nodes removed \\
Surg.oth.treat & Other surgical treatments performed \\
Surg.Rad & Surgery plus radiotherapy indicator \\
Radiation & Radiotherapy received (yes/no) \\
\hline
\multicolumn{2}{l}{\textbf{Survival outcomes and follow-up}} \\
OS\_Code / CSS\_Code & Overall survival and cause-specific survival event codes \\
Survival.months & Survival duration in months \\
COD.to.site.recode & Cause of death / vital status \\
Year.of.diagnosis & Year of initial diagnosis \\
\bottomrule
\end{tabular}
\end{table}

\begin{table}[htbp]
\setlength{\tabcolsep}{3pt} 
\renewcommand{\arraystretch}{1.25}
\centering
\caption{Structure of the generated electronic medical record and field definitions.}
\label{tab:emr_fields_grouped}
\begin{tabular}{p{6cm} p{8cm}}
\hline
\textbf{Field} & \textbf{Description} \\
\hline
\multicolumn{2}{l}{\textbf{General information}} \\
Sex & Patient's biological sex \\
Age & Age at initial diagnosis \\
PS score & Eastern Cooperative Oncology Group performance status (ECOG PS) \\
Extrahepatic\_metastasis & Indicator of metastasis beyond the liver (Yes/No) \\
\hline
\multicolumn{2}{l}{\textbf{Child\_Pugh score}} \\
Score & Overall Child--Pugh score (A (0), B (1), or C (2)) \\
Total\_bilirubin & Total bilirubin level \\
Serum\_albumin & Serum albumin concentration \\
Prothrombin\_time & Prolongation of prothrombin time (seconds) \\
Ascites & Ascites status (none, mild, severe) \\
Hepatic\_encelopathy & Presence and grade of hepatic encephalopathy \\
\hline
\multicolumn{2}{l}{\textbf{Examination related}} \\
Laboratory\_examination & Structured laboratory test results (blood counts, liver function, tumor markers, viral serologies) \\
Pathological\_examination & Biopsy-based pathology reports (gross description, microscopic features, diagnostic conclusions) \\
Imaging\_examination & Radiology examination reports (e.g., abdominal CT, chest CT) including findings and impressions \\
\hline
\end{tabular}
\end{table}

\begin{landscape}
  \captionof{table}{Baseline characteristics of the external multi-center HCC cohort}
  \label{tab:external_multicenter_baseline}
\begin{tiny}
\setlength{\tabcolsep}{3pt}
\renewcommand{\arraystretch}{1.12}
\begin{tabularx}{\linewidth}{p{2.0cm}|X|X|X|X|X|X|X|X|X|X|X|X}
\toprule
Characteristics & Center A      (N=1398)      & Center B      (N=1994)      & Center C      (N=945)       & Center D      (N=865)       & Center E      (N=431)       & Center F      (N=319)       & Center G      (N=299)       & Center H      (N=165)       & Center I       (N=76)       & Center J       (N=63)       & Center K       (N=57)       & Center L       (N=56)       \\ \midrule
\midrule
\endfirsthead
\multicolumn{13}{l}{\small\emph{Continued on next page}} \\
\toprule
Characteristics & Center A      (N=1398)      & Center B      (N=1994)      & Center C      (N=945)       & Center D      (N=865)       & Center E      (N=431)       & Center F      (N=319)       & Center G      (N=299)       & Center H      (N=165)       & Center I       (N=76)       & Center J       (N=63)       & Center K       (N=57)       & Center L       (N=56)       \\ \midrule
\midrule
\endhead
\midrule \multicolumn{13}{r}{\small\emph{Continued on next page}} \\
\endfoot
\bottomrule
\endlastfoot
\rowcolor{gray!30}
\textbf{Sex}  &  &  &  &  &  &  &  &  &  &  &  &  \\
\rowcolor{gray!10}
    female &   260 (18.6\%)    &   406 (20.4\%)    &   213 (22.5\%)    &   109 (12.6\%)    &    93 (21.6\%)    &    55 (17.2\%)    &    47 (15.7\%)    &    29 (17.6\%)    &    10 (13.2\%)    &    9 (14.3\%)     &    9 (15.8\%)     &    8 (14.3\%)     \\
\rowcolor{gray!10}
    male &   1138 (81.4\%)   &   1588 (79.6\%)   &   732 (77.5\%)    &   756 (87.4\%)    &   338 (78.4\%)    &   264 (82.8\%)    &   252 (84.3\%)    &   136 (82.4\%)    &    66 (86.8\%)    &    54 (85.7\%)    &    48 (84.2\%)    &    48 (85.7\%)    \\
\rowcolor{gray!10}
Age (years) & 59.0 [51.0;66.0] & 60.0 [53.0;66.0] & 59.0 [50.0;66.0] & 57.0 [48.0;64.0] & 61.0 [53.0;67.0] & 61.0 [53.0;68.0] & 56.0 [47.0;65.0] & 52.0 [46.0;60.0] & 61.0 [51.0;69.0] & 54.0 [45.0;62.0] & 60.0 [52.0;65.0] & 54.5 [45.8;65.2] \\
\rowcolor{gray!10}
BMI ($kg/m^2$) & 23.8 [21.5;25.8] & 24.2 [22.3;26.8] & 22.8 [20.8;24.8] & 23.7 [21.7;26.3] & 23.6 [21.4;26.0] & 23.5 [20.4;26.7] & 23.5 [21.3;25.6] & 23.4 [21.3;25.6] & 23.9 [22.4;26.1] & 23.5 [20.8;25.4] & 23.7 [21.2;26.8] & 24.3 [20.4;26.0] \\
\rowcolor{gray!30}
\textbf{Diabetes}  &  &  &  &  &  &  &  &  &  &  &  &  \\
\rowcolor{gray!10}
    No &   727 (73.7\%)    &   1603 (80.4\%)   &   696 (85.1\%)    &   682 (82.8\%)    &   286 (88.5\%)    &   254 (79.6\%)    &   252 (84.3\%)    &   136 (82.4\%)    &    57 (75.0\%)    &    56 (88.9\%)    &    40 (70.2\%)    &    51 (91.1\%)    \\
\rowcolor{gray!10}
    Yes &   260 (26.3\%)    &   391 (19.6\%)    &   122 (14.9\%)    &   142 (17.2\%)    &    37 (11.5\%)    &    65 (20.4\%)    &    47 (15.7\%)    &    29 (17.6\%)    &    19 (25.0\%)    &    7 (11.1\%)     &    17 (29.8\%)    &    5 (8.93\%)     \\
\end{tabularx}
\end{tiny}
\end{landscape}
\addtocounter{table}{-1}
\begin{landscape}
  \captionof{table}{Baseline characteristics of the external multi-center HCC cohort (survival-filtered testing set).}
  \label{tab:external_multicenter_baseline_filtered}
\begin{tiny}
\setlength{\tabcolsep}{3pt}
\renewcommand{\arraystretch}{1.12}
\begin{tabularx}{\linewidth}{p{4.5cm}|X|X|X|X|X|X|X|X|X|X|X|X}
\toprule
Characteristics & Center A      N=412       & Center B      N=1941      & Center C       N=61       & Center D      N=817       & Center E      N=143       & Center F      N=314       & Center G      N=249       & Center H       N=47       & Center I       N=76       & Center J       N=18       & Center K       N=56       & Center L       N=56       \\ \midrule
\midrule
\endfirsthead
\multicolumn{13}{l}{\small\emph{Continued on next page}} \\
\toprule
Characteristics & Center A      N=412       & Center B      N=1941      & Center C       N=61       & Center D      N=817       & Center E      N=143       & Center F      N=314       & Center G      N=249       & Center H       N=47       & Center I       N=76       & Center J       N=18       & Center K       N=56       & Center L       N=56       \\ \midrule
\midrule
\endhead
\midrule \multicolumn{13}{r}{\small\emph{Continued on next page}} \\
\endfoot
\bottomrule
\endlastfoot
\rowcolor{gray!30}
\textbf{Sex }  &  &  &  &  &  &  &  &  &  &  &  &  \\
\rowcolor{gray!10}
    female &    55 (13.3\%)    &   397 (20.5\%)    &    11 (18.0\%)    &   104 (12.7\%)    &    33 (23.1\%)    &    55 (17.5\%)    &    40 (16.1\%)    &    11 (23.4\%)    &    10 (13.2\%)    &    2 (11.1\%)     &    9 (16.1\%)     &    8 (14.3\%)     \\
\rowcolor{gray!10}
    male &   357 (86.7\%)    &   1544 (79.5\%)   &    50 (82.0\%)    &   713 (87.3\%)    &   110 (76.9\%)    &   259 (82.5\%)    &   209 (83.9\%)    &    36 (76.6\%)    &    66 (86.8\%)    &    16 (88.9\%)    &    47 (83.9\%)    &    48 (85.7\%)    \\
\rowcolor{gray!10}
Age (years) & 56.5 [50.0;63.0] & 60.0 [53.0;66.0] & 57.0 [49.0;62.0] & 57.0 [48.0;64.0] & 60.0 [53.0;67.0] & 61.0 [53.0;68.0] & 57.0 [48.0;66.0] & 52.0 [41.5;59.0] & 61.0 [51.0;69.0] & 52.5 [45.0;60.0] & 60.0 [51.5;65.0] & 54.5 [45.8;65.2] \\
\rowcolor{gray!10}
BMI (kg/m2) & 23.8 [21.5;26.1] & 24.2 [22.3;26.8] & 22.6 [20.9;24.0] & 23.7 [21.7;26.3] & 23.5 [21.4;25.9] & 23.5 [20.3;26.8] & 23.7 [21.3;25.8] & 23.5 [20.7;25.6] & 23.9 [22.4;26.1] & 23.7 [22.1;25.6] & 24.0 [21.2;26.8] & 24.3 [20.4;26.0] \\
\rowcolor{gray!30}
\textbf{Diabetes }  &  &  &  &  &  &  &  &  &  &  &  &  \\
\rowcolor{gray!10}
    No &   271 (70.0\%)    &   1564 (80.6\%)   &    52 (85.2\%)    &   646 (82.7\%)    &   104 (89.7\%)    &   252 (80.3\%)    &   211 (84.7\%)    &    41 (87.2\%)    &    57 (75.0\%)    &    15 (83.3\%)    &    39 (69.6\%)    &    51 (91.1\%)    \\
\rowcolor{gray!10}
    Yes &   116 (30.0\%)    &   377 (19.4\%)    &    9 (14.8\%)     &   135 (17.3\%)    &    12 (10.3\%)    &    62 (19.7\%)    &    38 (15.3\%)    &    6 (12.8\%)     &    19 (25.0\%)    &    3 (16.7\%)     &    17 (30.4\%)    &    5 (8.93\%)     \\
\end{tabularx}
\end{tiny}
\end{landscape}



\begin{appendices}
\clearpage


\setcounter{page}{1} 
\section{Objective Physician Scoring Questionnaire (CoT)}
\label{sec:treat_eval}

\noindent \textbf{Scope.} This consensus-based questionnaire, developed by senior hepatobiliary specialists, evaluates AI-generated content at the \emph{treatment} level (resection, ablation, transplant referral, TACE, radiotherapy, systemic therapy, clinical trial, supportive/palliative care, defer-and-reassess). It does not assess specific drugs or dosing.

\subsection{Basic Information}
\noindent Case ID:\ \rule{2.8cm}{0.4pt}\quad
Rater ID/Name:\ \rule{3.6cm}{0.4pt}\quad
Specialty:\ \rule{2.8cm}{0.4pt}\quad
Seniority (Resident / Attending / Senior Consultant):\ \rule{3.2cm}{0.4pt}

\subsection{Completeness (tick each if covered)}
\addtocounter{table}{-1}
\begin{table}[h]
\centering
\caption{Coverage checklist of required items}
\begin{tabular}{p{10.0cm} p{3.0cm}}
\toprule
Item & Choice \\
\midrule
Staging reference provided (at least one of BCLC/CNLC/TNM) & $\square$ Yes\quad $\square$ No \\
ECOG performance status is stated & $\square$ Yes\quad $\square$ No \\
Child--Pugh score is stated & $\square$ Yes\quad $\square$ No \\
Tumor burden considered (number and largest diameter) & $\square$ Yes\quad $\square$ No \\
Vascular invasion assessed (incl.\ portal vein tumor thrombus grading) & $\square$ Yes\quad $\square$ No \\
Extrahepatic disease / nodal status assessed & $\square$ Yes\quad $\square$ No \\
Portal hypertension / bleeding risk or FLR adequacy assessed & $\square$ Yes\quad $\square$ No \\
Transplant eligibility and contraindications assessed (incl.\ bridge/downstaging) & $\square$ Yes\quad $\square$ No \\
\bottomrule
\end{tabular}
\label{tab:completeness}
\end{table}

\noindent \textbf{Completeness score (0--40):}\ \rule{1.6cm}{0.4pt}\quad
\emph{Mapping:} each item with five points.

\subsection{Correctness (0--25; deduction rule)}

\noindent \textbf{Count of errors (subtract per item):}\quad
Major (each $-2$):\ $\square$ 0\quad $\square$ 1\quad $\square$ 2\quad $\square$ $\ge$3\qquad
Minor (each $-1$):\ $\square$ 0\quad $\square$ 1\quad $\square$ 2\quad $\square$ $\ge$3

\medskip
\noindent \textbf{Examples of major errors:}
\begin{itemize}\setlength\itemsep{2pt}
  \item Incorrect clinical staging assignment (BCLC/CNLC/TNM).
\end{itemize}

\noindent \textbf{Examples of minor errors:}
\begin{enumerate}\setlength\itemsep{2pt}
  \item Recommended option inconsistent with stage-appropriate guideline (e.g., stage-specific CNLC recommendations).
  \item “Best option” selection is inappropriate given the stage and eligibility.
  \item Recommendation appears guideline-concordant on the surface but violates activating conditions or contraindications for that option.
\end{enumerate}

\noindent \textbf{Final correctness score} $= 25 - 2\times(\text{major}) - 1\times(\text{minor})$:\ \rule{1.8cm}{0.4pt}

\subsection{Safety and Red Flags}

\noindent \textbf{Any red flag present? (any ``Yes'' $\Rightarrow$ Safety-Fail):}\quad
$\square$ Yes ($<2$ red flags)\ \ \ $\square$ No ($>=2$ red flags)

\begin{table}[h]
\centering
\caption{Red-flag checklist (any ``Yes'' $\Rightarrow$ Safety-Fail)}
\begin{tabular}{p{10.0cm} p{3.0cm}}
\toprule
Red-flag item & Choice \\
\midrule
Child--Pugh C or ECOG $>$ 2 yet recommending surgery/intervention/systemic therapy & $\square$ Yes\quad $\square$ No \\
Severe infection, organ failure, uncorrectable coagulopathy, or active hepatitis yet recommending surgery/intervention & $\square$ Yes\quad $\square$ No \\
Main-trunk portal vein complete thrombosis with TACE proposed as routine & $\square$ Yes\quad $\square$ No \\
Widespread extrahepatic disease yet curative local therapy prioritized (resection/ablation) & $\square$ Yes\quad $\square$ No \\
Marked portal hypertension or inadequate FLR yet resection prioritized & $\square$ Yes\quad $\square$ No \\
\bottomrule
\end{tabular}
\label{tab:safety}
\end{table}

\noindent \textbf{Safety score:}\quad 25 - $5\times(\text{red flag})$



\subsection{Summary and Adoption Decision}

\begin{table}[h]
\centering
\caption{Domain scores, composite, and adoption decision}
\begin{tabular}{p{8.0cm} p{5.5cm}}
\toprule
Item & Entry \\
\midrule
Completeness (0--40) & \rule{3cm}{0.4pt} \\
Correctness (0--25) & \rule{3cm}{0.4pt} \\
Safety (0--25) & \rule{3cm}{0.4pt} \\
\midrule
Core composite QNS (sum of Completeness, Correctness, and Safety) & \rule{3cm}{0.4pt}\, \\
Safety-Pass (no red flag and Safety $\ge$ 3) & $\square$ Yes\quad $\square$ No \\
Adoption decision & $\square$ Adopt\quad $\square$ Adopt with edits\quad $\square$ Do not adopt \\
\bottomrule
\end{tabular}
\end{table}

\subsection*{Scoring Notes}
\begin{enumerate}\setlength\itemsep{4pt}
  \item The questionnaire assesses \emph{treatment-category} recommendations only; it does not score specific drugs or dosing.
  \item Completeness is mapped from the count of required items covered.
  \item Correctness uses a deduction rule: start at 25; subtract 2 per major error and 1 per minor error; floor at 0.
  \item Safety uses a red-flag mechanism: any red flag $\Rightarrow$ Safety-Fail and the number of red flags is less than 2 $\Rightarrow$ Safety-Pass.
  \item Composite and adoption thresholds: QNS $\ge 80$ and Safety-Pass $\Rightarrow$ \emph{Adopt}; $70 \le$ QNS $< 80$ and Safety-Pass $\Rightarrow$ \emph{Adopt with edits}; otherwise \emph{Do not adopt}.
\end{enumerate}

\section{Scoring rubric for evidence-based justification}
\label{sec:score_evidence}
\noindent Please rate each of the three decision processes along the following three evidence-based dimensions. Each dimension is scored out of 10 points.
\begin{enumerate}
  \item \textbf{Completeness of cited evidence} (whether every recommended option is accompanied by an evidence citation).\\
  \textbf{10 points}: every recommended option has an accompanying evidence citation; \textbf{5 points}: most options have evidence; \textbf{3 points}: only a minority of options have evidence; \textbf{0 points}: no options have evidence.

  \item \textbf{Relevance to the index patient} (whether the cited evidence correctly corresponds to the current patient’s condition).\\
  \textbf{10 points}: all cited evidence is highly relevant to the current patient; \textbf{5 points}: most cited evidence is relevant; \textbf{3 points}: only a minority is relevant; \textbf{0 points}: none of the cited evidence is relevant.

  \item \textbf{Accuracy of guideline citation} (whether guideline sources are cited correctly).\\
  \textbf{10 points}: all guideline citations are correct; \textbf{5 points}: most are correct; \textbf{3 points}: only a minority are correct; \textbf{0 points}: none are correct.\\
  \emph{Example:} if a statement reads “For hepatocellular carcinoma with diameter $\leq 1$\,cm, overall survival with surgical resection is similar to or slightly better than ablation (evidence level 3, recommendation B),” but the cited evidence level does not exist in the guideline, this is counted as an incorrect citation.
\end{enumerate}

\clearpage

\renewcommand{\figurename}{Figure}
\renewcommand{\tablename}{Table}
\renewcommand{\thefigure}{A\arabic{figure}} 
\renewcommand{\thetable}{A\arabic{table}}
\setcounter{table}{1}

\setcounter{figure}{0}
\newpage
\label{secA1}
\begin{lstlisting}[style=promptmono,floatplacement=h,caption={The summary of decision tree of CNLC},label={lst:hcc_prompt_cnlc}]
# Instruction
Assume you are a senior hepatobiliary clinical researcher. Based on the patient's baseline and imaging/treatment information, rate suitability (0-1) for each option and explain step-by-step according to the CNLC-style rules below.

# Options
1) Surgical resection
2) Ablation
3) Interventional therapy (e.g., TACE)
...

# Rules (concise; CNLC decision tree, English version)

1) Baseline screening
   - Performance Status (PS) and Child-Pugh:
     * If PS 3-4 OR Child-Pugh C -> classify as Stage IV (terminal set): supportive care / liver transplantation / palliative care.
     * If PS 0-2 AND Child-Pugh A or B -> continue to Step 2.

2) Tumor burden staging (precondition: no macrovascular invasion AND no extrahepatic metastasis)
   - Single tumor:
     * Size <= 5 cm -> Stage Ia.
       Tx: surgical resection / ablation / liver transplantation.
     * Size > 5 cm -> Stage Ib.
       Tx: surgical resection / interventional therapy / ablation / interventional therapy + ablation / consider liver transplantation.
   - 2-3 tumors:
     * All <= 3 cm -> Stage IIa.
       Tx: surgical resection / interventional therapy / resection + ablation / interventional therapy + ablation / liver transplantation.
     * Any > 3 cm -> Stage IIb.
       Tx: interventional therapy / surgical resection / systemic anti-tumor therapy.
   - >= 4 tumors -> Stage IIb.
       Tx: interventional therapy / surgical resection / systemic anti-tumor therapy.

3) Invasion and metastasis assessment
   - Vascular invasion present and no distant/extrahepatic metastasis -> Stage IIIa.
     Tx: interventional therapy + systemic anti-tumor therapy / surgical resection / radiotherapy.
   - Extrahepatic metastasis present -> Stage IIIb.
     Tx: systemic anti-tumor therapy / interventional therapy / radiotherapy.

4) Terminal criteria re-check
   - If PS 3-4 OR Child-Pugh C at any point -> Stage IV (terminal set): supportive care / liver transplantation / palliative care.
...
\end{lstlisting}

\begin{lstlisting}[style=promptmono,floatplacement=H,caption={An template of synthetic electronic medical record in JSON format},label={lst:json_emrs}]
{
  "generated_electronic_medical_record": {
    "PS": 2,
    "Child_Pugh": {
      "score": "B",
      "total_bilirubin": 42,
      "serum_albumin": 3.1,
      ... 
    },
    "extrahepatic_metastasis": "Yes",
    "examination_related": {
      "laboratory_examination": {
        "white_blood_cell_count": 6.2,
        "red_blood_cell_count": 4.3,
        "hemoglobin": 125,
        ... 
      },
      "pathological_examination": {
        "biopsy_pathological_diagnosis": {
          "gross_description": "Liver biopsy: several gray-white tissue strips, 2.5-3cm long, 0.2cm diameter, intact.",
          "preliminary_diagnosis": "(Liver lesion) Moderately differentiated hepatocellular carcinoma with vascular invasion; ...",
          ... 
        }
      },
      "imaging_examination": {
        "upper_abdominal_CT_scan": {
          "imaging_impression": "Right liver lobe lesion consistent with hepatocellular carcinoma, with necrosis and vascular invasion; ...",
          ... 
        },
        ... 
      }
    }
  }
}
\end{lstlisting} 

\begin{lstlisting}[style=promptmono,floatplacement=H,caption={EMR-generation prompt},label={lst:emr_gen_prompt}]
# Instruction
Assume that you are a senior clinical medical researcher, and you now need to complete the following tasks:
You will be given relevant information about a patient, including:
1. Some basic information about the patient
2. Some laboratory test results of the patient before treatment
3. Some indicators of imaging examinations before treatment
4. The patient's current staging, such as T N M
What you need to give is to generate an electronic medical record (EMR) that looks real and conforms to the patient's situation based on this information...
The given patient information may not directly contain all necessary EMR content. You need to generate a simulated EMR that does not contradict existing data and output your reasoning process.

# The generated EMR should include:
1. PS: Patient's physical activity status (0 = fully normal, 4 = bedridden)
2. Child-Pugh_score: based on total bilirubin, serum albumin, prothrombin time, ascites, hepatic encephalopathy.
   - Grades: A: 5-6 points, B: 7-9 points, C: 10-15 points.
   - Example thresholds:
     A) Total bilirubin (umol/L): 1pt <34; 2pt 34-50; 3pt >50
     B) Serum albumin (g/dL): 1pt >3.5; 2pt 2.8-3.5; 3pt <2.8
     c) ...
3. extrahepatic_metastasis: ``Yes'' or ``No''.
4. examination_related: include key parts related to laboratory, pathology, and imaging examinations.
...

# Thinking process
Provide a detailed step-by-step explanation of how you constructed the EMR, including reasoning and assumptions.
Example (truncated):
"The patient is 90 years old, with a large tumor and severe condition, thus PS = 3.
Assume total bilirubin = 35 (34-50), serum albumin = 3.1 (2.8-3.5),
prothrombin time = 4.5 (4.0-6.0), no ascites, no encephalopathy.
Calculated Child-Pugh = 8, corresponding to Grade B.
No extrahepatic metastasis (mets.x = 0).
AFP is elevated (421 ng/mL), consistent with imaging findings..."
...

# Output format (JSON only)
{
  "thinking": "<your-step-by-step-reasoning>",
  "generated_electronic_medical_record": "<Your-generated-medical-record>"
}

# Hint
1. Some information may be missing (originally shown as 'Blank(s)').
2. Output only the JSON object above (no extra text).
...

# Example (truncated)
{"PS": 1, "Child_Pugh": {"score": "A", "total_bilirubin": 30, "serum_albumin": 3.8, ...},
 "extrahepatic_metastasis": "No",
 "examination_related": {"laboratory_examination": {...}, "pathological_examination": {...}, ...}}
...
\end{lstlisting}

\begin{lstlisting}[style=promptmono,floatplacement=H,caption={Clinical-Reasoning prompt for generating thinking data},label={lst:hcc_prompt_thinking}]
# Instruction
Imagine you are a senior hepatobiliary surgeon. Given the patient baseline, labs, and imaging, produce the internal "thinking" before the final recommendation. Rate each option (0-1) and justify.

# Inputs (may include)
- Basic info, comorbidities, hepatitis, prior therapy, ...
- Laboratory tests (ALB, TBIL, INR, AFP, ...), Child-Pugh
- Imaging reports (tumor number/size, segment, VI, EHM, ...)

The treatment options you are presented with include:
1) Surgical resection
2) Ablation
...

# Tasks: In addition to assigning a recommendation score to each treatment option, you are required to provide your step-by-step reasoning. 
1) Extract key evidence from input (PS, Child-Pugh, tumor number/size, VI, EHM, ...).
2) Locate the branch in CNLC/BCLC/TNM according to the previously provided decision trees (...).
3) From the CNLC branch, list all guideline-suggested alternatives for this stage (...).
4) For each alternative, assess: indications met, contraindications not violated, missing-but-needed info, reasonable inference if any, evidence level and recommendation grade.
5) Compare alternatives and rank; assign scores 0-1 with short reasons.

# Checking rules
- The "preliminary decision by CNLC" must include all options suggested by that CNLC stage.
- The "detailed analysis" must cover every option listed above.
- CNLC, BCLC, TNM determinations must be consistent with the decision trees shown earlier.

# Background knowledge 
- BCLC tree summary: ...
- CNLC tree summary: ...

# Output format (JSON schema, parsable by json.loads)
{
  "thinking": "<step-by-step natural-language reasoning in Chinese ...>",
  "check_for_thinking": "<boolean or short note that all CNLC-suggested options were covered ...>",
  "scores": {
    "Surgical_resection": {"Indications": <0-1>, "Not_Violate_Taboos": <0-1>, "Level_of_Evidence": <0-1 or -1>, "Level_of_Recommendation": <0-1 or -1>, "Comprehensive_Score": <0-1>}}
}

# Example
Here are some good examples of output:
Thinking:
- Step 1: Baseline -> PS 0-2 and Child-Pugh A/B -> proceed ...
- Step 2: From CNLC tree -> single tumor <= 5 cm -> candidates: resection / ablation / transplantation ...
- Step 3: Compare: resection favored due to ...; ablation acceptable given ...; transplantation not prioritized due to ...
- Ranking: resection > ablation > transplantation ...
- Scores: {"Surgical_resection": {"Comprehensive_Score": 0.9}, "Ablation": {"Comprehensive_Score": 0.7}, ...}

# Constraints and hints
1) Do not mention or imply knowledge of real treatment or survival outcomes.
2) Do not include phrases implying prior knowledge of survival months.
3) ...
\end{lstlisting}

\begin{lstlisting}[style=promptmono,floatplacement=H,caption={Tagging prompt (the complete prompt is provided in Supplementary Materials)},label={lst:tag_en_full}]
# Instruction
Suppose you are a clinical researcher. Now you will be given a reasoning process for treatment decision. What you need to do is to add labels before and after the key data. The specific labels and corresponding data content examples are as follows:

A. PS score: <ps>1</ps>
B. Child-Pugh grade: <child_pugh>A</child_pugh>
C. Extrahepatic metastasis: <metastasis>...</metastasis>   // allowed values: 'present' or 'absent'
D. Imaging-visible cancer thrombus: <cancer_thrombus>...</cancer_thrombus>   // allowed values: 'present' or 'absent'
E. Number of tumors: <num_tumor>1</num_tumor>
F. Maximum tumor size (mm): <tumor_size>30</tumor_size>
G. CNLC staging: <cnlc></cnlc>, BCLC staging: <bclc></bclc> and TNM staging: <tnm></tnm> 
H. Suggested treatment:<treatment>...</treatment>, where the optional treatments include:
1) Surgical resection
2) Ablation
...
I. Treatment not suggested: <treatment_not_recommended>...</treatment_not_recommended>

Here is an excellent example (excerpts):
"The patient's performance status is <ps>2</ps>, and the Child-Pugh grade is <child_pugh>B</child_pugh>, indicating an overall condition that allows further evaluation rather than defaulting to supportive or palliative care. Imaging shows <metastasis>present</metastasis> extrahepatic metastasis, and the maximum tumor diameter is <tumor_size>53</tumor_size> mm. There is <cancer_thrombus>absent</cancer_thrombus> vascular invasion on imaging. Guidelines-staging: <cnlc>II</cnlc>. Considering the tumor burden and dissemination, we recommend <treatment>Symptomatic support</treatment>, an interventional option such as <treatment>TACE</treatment>, or <treatment>Radiotherapy</treatment>, and we do not recommend <treatment_not_recommended>Surgical resection</treatment_not_recommended>."

# Input
'{reasoning}'

# Hint
1) Your final output must be in Chinese and must consist only of the labeled reasoning text (no extra commentary).
2) If a specific item is missing in the passage, skip that tag.
3) The <metastasis>...</metastasis> tag is ONLY for extrahepatic metastasis (do not use it for other metastases).

...
\end{lstlisting}

\end{appendices}


\end{document}